\PassOptionsToPackage{table}{xcolor}
\documentclass[runningheads]{llncs}

\usepackage{eccv}

\usepackage{eccvabbrv}
\usepackage{graphicx}
\usepackage{booktabs}
\usepackage[accsupp]{axessibility} % Improves PDF readability
\usepackage{orcidlink}

\usepackage[most]{tcolorbox}
\usepackage{tabularx}
\usepackage{multirow}
\usepackage{siunitx}
\usepackage{tikz}
\usepackage{float}
\usepackage{titletoc}
\usepackage{hyperref}
\usepackage{marvosym}

\definecolor{eccvblue}{rgb}{0.21,0.49,0.74}
\definecolor{bev_yellow}{RGB}{250,186,60}
\definecolor{bev_green}{RGB}{224,240,227}
\definecolor{bev_blue}{RGB}{76,132,239}

\newcommand{\colorblock}[2]{%
  \begin{tcolorbox}[
    colback=#1, colframe=#1,
    arc=3mm, boxrule=0pt,
    left=4pt, right=4pt, top=2pt, bottom=2pt,
    width=\linewidth,
    enlarge left by=0mm, enlarge right by=0mm
  ]
  #2
  \end{tcolorbox}
}

\newcommand\blfootnote[1]{%
  \begingroup
  \renewcommand\thefootnote{}{}\footnote{#1}%
  \addtocounter{footnote}{-1}%
  \endgroup
}

\makeatletter
\def\blfootnote{\gdef\@thefnmark{}\@footnotetext}
\makeatother

\newtcbox{\answerbox}[1][]{%
    enhanced, nobeforeafter, tcbox raise base,
    boxrule=0.5mm,
    colback=bev_yellow!10, colframe=bev_yellow!,
    coltext=bev_yellow!40!black,
    fontupper=\bfseries,
    arc=4mm, boxsep=1mm,
    left=3mm, right=3mm, top=2mm, bottom=2mm,
    #1
}

\newtcolorbox{questionbox}{%
    enhanced,
    colback=bev_blue!8, colframe=bev_blue!,
    coltext=bev_blue!30!black,
    boxrule=0.5mm, arc=4mm, boxsep=1mm,
    left=2mm, right=2mm, top=2mm, bottom=2mm,
    width=\linewidth,
}

\newcommand{\qaside}[2]{%
    \noindent
    \renewcommand\tabularxcolumn[1]{m{##1}}
    \begin{tabularx}{\linewidth}{@{} X @{\hspace{3mm}} c @{}}
        \begin{questionbox}
        \textbf{Q:} #1
        \end{questionbox}
        &
        \answerbox{A: #2} \\
    \end{tabularx}
}

\providecommand{\authcount}[1]{}

\title{BEVLM: Distilling Semantic Knowledge from LLMs into Bird's-Eye View Representations}

\titlerunning{BEVLM}

\author{Thomas Monninger\inst{1,*} \and
Shaoyuan Xie\inst{2,*,\dagger} \and
Qi Alfred Chen\inst{2} \and
Sihao Ding\inst{1,}\textsuperscript{\Letter}}

\authorrunning{T. Monninger, S. Xie, Q. A. Chen, and S. Ding}

\institute{Mercedes-Benz Research \& Development North America, San Jose, USA \and
University of California, Irvine, USA\\
\email{\{thomas.monninger, sihao.ding\}@mercedes-benz.com}\\
\email{\{shaoyux, alfchen\}@uci.edu}}

\begin{document}
\maketitle

\setcounter{tocdepth}{3}

% Footnote
\blfootnote{$(*)$ Equal contribution, the order was determined alphabetically.}
\blfootnote{$(\dagger)$ Work was done during an internship at Mercedes-Benz Research \&
Development North America.}
\blfootnote{${(\textrm{\Letter})}$ Corresponding author.}

\begin{abstract}
The integration of Large Language Models (LLMs) into autonomous driving has attracted growing interest for their strong reasoning and semantic understanding abilities, which are essential for handling complex decision-making and long-tail scenarios. However, existing methods typically feed LLMs with tokens from multi-view and multi-frame images independently, leading to redundant computation and limited spatial consistency. This separation in visual processing hinders accurate 3D spatial reasoning and fails to maintain geometric coherence across views. On the other hand, Bird's-Eye View (BEV) representations learned from geometrically annotated tasks (\eg, object detection) provide spatial structure but lack the semantic richness of foundation vision encoders. To bridge this gap, we propose \textbf{BEVLM}, a framework that connects a spatially consistent and semantically distilled BEV representation with LLMs. Through extensive experiments, we show that BEVLM enables LLMs to reason more effectively in cross-view driving scenes, improving accuracy by \SI{46.0}{\percent}, by leveraging BEV features as unified inputs. Furthermore, by distilling semantic knowledge from LLMs into BEV representations, BEVLM significantly improves closed-loop end-to-end driving performance in safety-critical scenarios across UniAD and VAD, with gains of up to \SI{28.2}{\percent}.
\end{abstract}
\section{Introduction}

\label{sec:intro}
Large Language Models (LLMs) have rapidly advanced in scene understanding and reasoning~\cite{yang2025qwen3, achiam2023gpt, comanici2025gemini, zhu2025internvl3}, attracting growing interest for autonomous driving applications~\cite{sima2023drivelm, jiang2024senna, pan2024vlp, xu2024vlm, wang2024omnidrive, xie2025drivebench}. Integrating language foundation models into autonomous driving systems provides a path towards commonsense reasoning, open-world understanding, and enhanced interpretability, capabilities often lacking in conventional perception or end-to-end driving pipelines~\cite{yang2023llm4drive}. Such reasoning ability is particularly critical for handling complex driving scenarios and corner cases, \ie, the ``long tail'' of the distribution.

\begin{figure}[t]
    \centering
    \includegraphics[width=\linewidth]{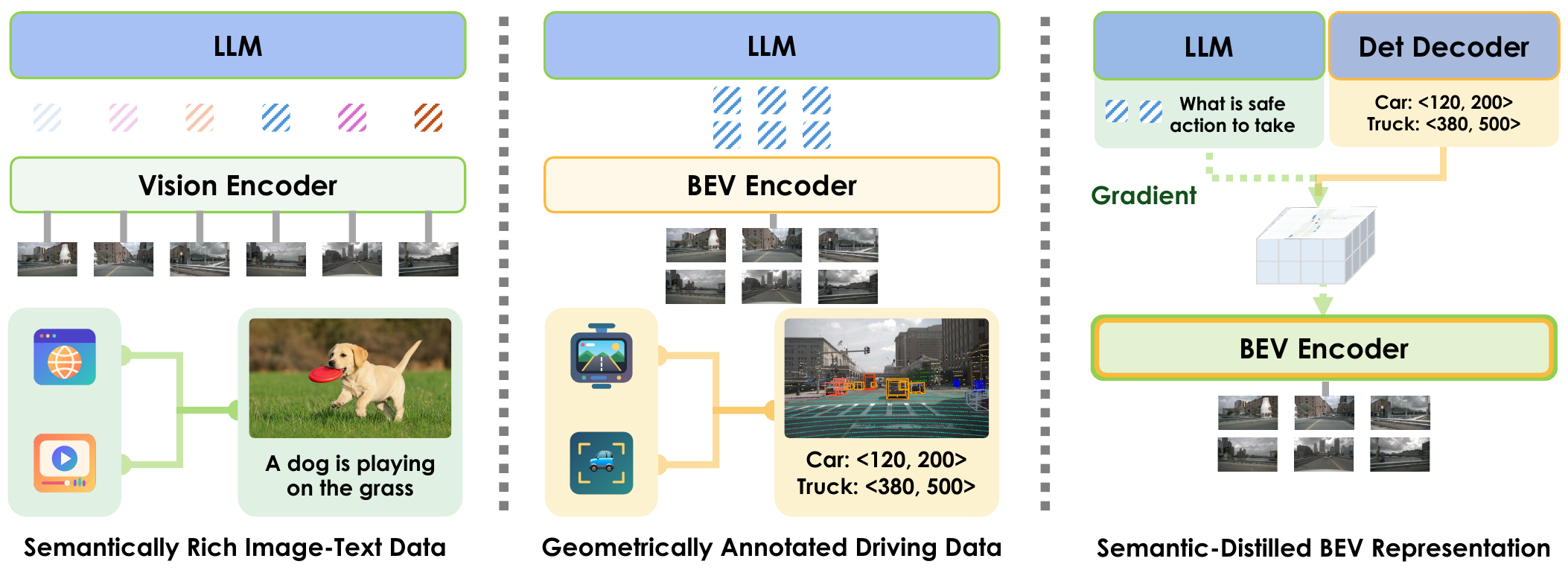}
    \caption{\textbf{Representation Comparison}: 
    Left: Vision encoders can leverage widely available semantically rich image-text data, but process multi-view images independently. Center: Bird's-Eye View (BEV) encoders provide a spatially consistent scene representation, but are limited to geometrically annotated data. Right (ours): We propose the semantic distillation from LLMs to BEV encoders to build a semantic-enhanced and spatially consistent scene representation.
    }
    % \vspace{-0.3cm}
    \label{fig:teaser}
\end{figure}

To integrate LLMs into autonomous driving, most existing systems leverage Vision Language Models (VLMs) and extract visual tokens independently from multi-view and multi-frame images~\cite{sima2023drivelm, jiang2024senna, tian2024drivevlm, shao2024lmdrive, zhou2025autovla, renz2025simlingo, zeng2025futuresightdrive, hwang2024emma, wang2025alpamayo}.
While this design is straightforward and leverages the large-scale pre-training of VLMs to align vision and language modalities, it introduces two key limitations. As shown in \cref{fig:teaser}, first, the resulting representations capture each view angle individually and are encoded into different token chunks. This separate processing fails to model spatial consistency, which is crucial for modeling dynamic driving environments~\cite{ma2024vision} and poses a challenge for 3D spatial reasoning of LLMs~\cite{gholami2025spatial}. Second, this design fails to exploit the temporal correlation, and separate processing makes the computational cost grow proportionally with the number of frames, leading to an inevitable trade-off between capturing long-term temporal information and maintaining computational efficiency~\cite{Zhou_2025_ICCV}.

Meanwhile, the Bird's-Eye View (BEV) representation has become a cornerstone of modern autonomous driving systems~\cite{ma2024vision}. BEV provides a unified top-down view of the 3D environment by fusing information from multiple viewpoints, time steps, and even sensor modalities into a compact and spatially consistent grid. This representation enables more effective reasoning about spatio-temporal relationships among the ego-vehicle, dynamic agents, and static surroundings, which is crucial for reliable scene understanding. Owing to these advantages, the BEV grid has become the de facto intermediate representation for object detection~\cite{li2022bevformer,liu2023bevfusion}, motion prediction~\cite{wu2020motionnet,zhang2022beverse}, and vehicle planning~\cite{hu2023planning,jiang2023vad}.

However, despite its compactness and spatial consistency, the BEV representation cannot be pre-trained at scale using semantically rich image–text datasets, as is possible with foundation visual encoders in VLMs~\cite{yang2025qwen3, achiam2023gpt, comanici2025gemini, zhu2025internvl3}. Large-scale pre-training is essential for learning transferable visual features that generalize to rare and open-world driving scenarios~\cite{radford2021learning}. The lack of such semantic richness forms a fundamental bottleneck, preventing BEV-based representations from being adopted by the advances of LLMs.

In this paper, we conduct the first rigorous experiments showing the advantages of the spatially consistent BEV representation for LLM reasoning in autonomous driving, which we call \textbf{Bird's-Eye View Language Model (BEVLM)}. BEV improves scene understanding accuracy by \SI{46.0}{\percent} over multi-view inputs and matches foundation vision encoders that are 10$\times$ larger, suggesting BEV is a superior scene representation for LLMs. Building on this, we use the BEVLM framework for semantic distillation, framing the LLM as a teacher that supervises the BEV encoder (student) via VQA tasks. This yields a semantic-aware BEV encoder that interacts effectively with language models while preserving spatial structure. Plugging it into two representative end-to-end driving frameworks, UniAD and VAD, both show consistent closed-loop gains in safety-critical scenarios (up to \SI{28.2}{\percent} higher safety score and \SI{16.1}{\percent} lower collision rate), confirming that the benefits generalize across BEV-based architectures. The project website is at~\url{https://sites.google.com/view/secure-safe-ai/bevlm}.

Our contributions are summarized as follows: 
\begin{enumerate}
    \item We are the first to carry out a representation study that compares individual multi-frame multi-view perspective images and joint BEV representations for LLM spatial reasoning in autonomous driving.
    \item We propose BEVLM, a framework to distill semantic information from LLMs into the BEV encoder while preserving the spatial BEV representation.
    \item We train an end-to-end driving model from the distilled BEV encoder and find significant improvements in closed-loop evaluation, confirming distillation performance specifically in safety-critical scenarios.
\end{enumerate}

\section{Related Work}
\label{sec:related-works}

\noindent
\textbf{LLMs for Autonomous Driving.}
With the rapid advancement of LLMs, there is a growing interest in applying their reasoning and knowledge capabilities to autonomous driving. The key motivation is to leverage the human knowledge and commonsense reasoning embedded in LLMs to better handle corner cases and long-tail scenarios~\cite{yang2023llm4drive}. Existing studies generally follow two directions: (1) using LLM-generated text as high-level guidance for BEV-based end-to-end driving pipelines~\cite{tian2024drivevlm, jiang2024senna, pan2024vlp, xu2024vlm, wang2024omnidrive, feng2025verdi}; and (2) directly generating driving trajectories through LLMs~\cite{sima2023drivelm, shao2024lmdrive, xu2024drivegpt4, hwang2024emma, chen2025drivinggpt, huang2024making, xie2025s4, zhou2025autovla, fu2025orion, zhou2025opendrivevla, guo2025vdt}. 
However, most of these approaches still follow the conventional VLM paradigm, where visual features are extracted independently from individual camera views and frames. This design limits the LLM’s ability to capture spatio-temporal consistency and geometric relationships across views. 
Recent works~\cite{winter2025bevdriver, brandstaetter2025bev, Zhou_2025_ICCV} begin to explore connecting BEV and language modalities. Yet, a systematic study comparing the representational advantages of image-based versus BEV-based inputs for LLM reasoning has been missing. Also, solutions for addressing the semantic gap between these two representations are still underexplored.

\noindent
\textbf{BEV Representation.}
The BEV representation combines and integrates information from multiple views, time steps, and even sensor modalities~\cite{ma2024vision, xie2024benchmarking}. It has become a central intermediate representation for full-stack autonomous driving, enabling perception~\cite{li2022bevformer, liu2023bevfusion, huang2021bevdet, zhang2022beverse, monninger2025mapdiffusion}, prediction~\cite{zhang2022beverse, hu2023planning, fadadu2022multi}, and planning~\cite{hu2023planning, jiang2023vad}. 
However, learning semantically rich BEV representations remains an open challenge~\cite{yang2024unipad, min2024driveworld}. Current BEV learning methods rely heavily on dense geometric supervision, often through object detection~\cite{li2022bevformer, liu2023bevfusion}, map construction~\cite{liao2023maptr, monninger2025augmapnet}, or joint end-to-end training~\cite{hu2023planning, jiang2023vad}. While such supervision provides strong geometric cues, it limits the semantic richness needed for understanding complex, safety-critical scenarios, which is an essential requirement for conditionally, highly, and fully automated driving systems.

\noindent
\textbf{Safety-Critical Evaluation.}
Autonomous driving is inherently a safety-critical task, where unsafe decisions can result in severe consequences~\cite{muhammad2020deep, wan2022too}. Most existing studies emphasize the robustness of the perception module, particularly under out-of-distribution inputs~\cite{xie2024benchmarking, kong2023robo3d} or adversarial perturbations~\cite{cao2021invisible, sato2021dirty, xie2024on}. In contrast, the safety of the planning module has received relatively limited attention. Recently, several benchmarks have been introduced to assess planning safety~\cite{ljungbergh2024neuroncap, dauner2024navsim, jia2024bench2drive}.
For instance, the NeuroNCAP benchmark~\cite{ljungbergh2024neuroncap} generates safety-critical driving scenarios through closed-loop simulation.
In this work, we focus on enhancing the semantic understanding of the BEV representation to promote safer decision-making in end-to-end autonomous driving systems.

\section{BEV Representation for Spatial Reasoning}
\label{sec:representation-study}

To understand how BEV representations affect spatial understanding, this section examines whether LLMs can effectively interpret and reason over BEV inputs, and then compares BEV and image-based representations in supporting spatial reasoning. The results of this study motivate \cref{sec:approach}, where we enhance BEV representations with semantic knowledge from LLMs.

\subsection{BEV-to-Language Alignment}
\label{sec:bev-tokenization}
\colorblock{gray!18}{\textit{\textbf{Research Question 1}: Can BEV representations be aligned with the language space such that LLMs can reason over them as effectively as task-specific BEV perception modules?}}

In this section, we examine whether BEV features can be fully aligned with the language space for LLM-based reasoning. We mainly compare two setups: (1) our BEVLM framework, which uses an LLM receiving BEV features through a learned projector~\cite{liu2023visual}, and (2) a task-specific BEV model whose outputs are converted to answers via rule-based logic. By comparing their accuracy in identifying objects in the scene, we can assess whether the BEV-to-language projection retains sufficient spatial information for the LLM to reason over as effectively as the task-specific decoder (e.g., detection head).

\begin{table}[t]
 \centering
 \caption{\textbf{BEV Projector Alignment Study.} We compare the performance between the BEV-based detected bounding box and the LLM that directly operates on the same BEV grid. We report binary classification accuracy. \textit{Majority class} denotes using the majority choices from the training set for each category. The final score is averaged across all 10 categories, as we only show the top 5 frequent objects. For full results, please refer to the Appendix~\ref{supp:more_experiment_results}.}
 \label{tab:mlp_projector}
 \resizebox{0.65\linewidth}{!}{
 \begin{tabular}{l|c|ccccc|c}
  \toprule
  \textbf{Model} & Modality & cars &  peds. & trucks & cones & barriers & \textbf{Avg.}
  \\
  \midrule
    Majority class & - & 92.7 & 81.9 & 65.0 & 59.6 & 54.8 & 78.2 \\ 
    Linear probe & - & 94.6 & 87.2 & 84.2 & 83.8 & 83.9 & 88.7 \\
    Detection$_\text{UniAD}$ & - & 91.1 & 90.9 & 86.7 & \textbf{99.0} & \textbf{99.6} & 92.8 \\
    \midrule
    \rowcolor{bev_blue!13} InternVL3$_\text{1B}$ & B$_\text{UniAD}$ & 94.7 & 88.9 & 85.8 & 90.1 & 88.6 & 90.8 \\
    \rowcolor{bev_yellow!13} InternVL3$_\text{8B}$ & B$_\text{UniAD}$ & 97.7 & \textbf{94.8} & \textbf{89.7} & 94.0 & 95.0 & \textbf{95.3} \\
    \midrule
    \rowcolor{bev_green!50} DeepSeek-VL$_\text{1B}$ & B$_\text{UniAD}$ & \textbf{95.1} & 92.0 & 84.9 & 92.6 & 92.2 & 92.2  \\
  \bottomrule
 \end{tabular}}
 \vspace{-0.3cm}
\end{table}

\noindent
\textbf{Experimental Setups.}
We assess whether the projector effectively aligns BEV features with the language space by comparing three results on binary \textit{object-existence} questions from DriveLM-nuScenes dataset~\cite{sima2023drivelm}. We train the projector on \textit{perception-related} question–answer pairs and evaluate on questions such as \textit{``Is there a \underline{\smash{moving car}} in the \underline{\smash{front left}}?''}. We adopt the BEV encoder from UniAD~\cite{hu2023planning} and the language component of InternVL3~\cite{zhu2025internvl3}. The BEV grid is max-pooled to $50\times50$, producing 2,500 tokens. The original vision encoder is disabled when BEV tokens are used. We choose the language component of a VLM over a pure text-based LLM since we anticipate better spatial understanding from the cross-modal alignment. Additionally, this setup enables us to conduct comparisons across different modalities (\cref{sec:comparative-study}). Data examples are provided in Appendix~\ref{sec:qa-example}. More implementation details can be found in Appendix~\ref{supp:more_implementation_details}.

\noindent
\textbf{Results.} \cref{tab:mlp_projector} reports both overall and class-wise accuracy for the five most frequent object categories. We compare BEVLM against three baselines: (1) a \textit{majority class} prior based on DriveLM-nuScenes training split~\cite{sima2023drivelm, xie2025drivebench}; (2) a stronger linear probe baseline, implemented as an object-specific linear classifier on max-pooled BEV features of the view specified in the question, to consider the influence of data bias on the learning process; and (3) UniAD’s detection head, where answers are derived by matching \textit{class}, \textit{moving state}, and \textit{view angle} to the detector outputs. BEVLM substantially outperforms the majority-class and linear probe baselines across all categories, with especially strong gains on more balanced classes such as trucks, cones, and barriers (near \SI{90}{\percent}). This indicates that the projector is not simply capturing dataset priors but preserving meaningful spatial and semantic cues for LLMs to answer correctly.

BEVLM also approaches the accuracy of UniAD’s task-specific decoder. Specifically, InternVL3$_\text{1B}$ achieves \SI{90.8}{\percent} and DeepSeek-VL$_\text{1B}$ achieves \SI{92.2}{\percent}, versus UniAD’s \SI{92.8}{\percent}. Scaling the InternVL3 to 8B yields \SI{95.3}{\percent}, surpassing the detector itself. These results show that a simple MLP projector can effectively map BEV features into the language space. Complete results are in Appendix~\ref{supp:more_experiment_results}.

\begin{figure}[t!]
    \centering
    \includegraphics[width=0.85\linewidth]{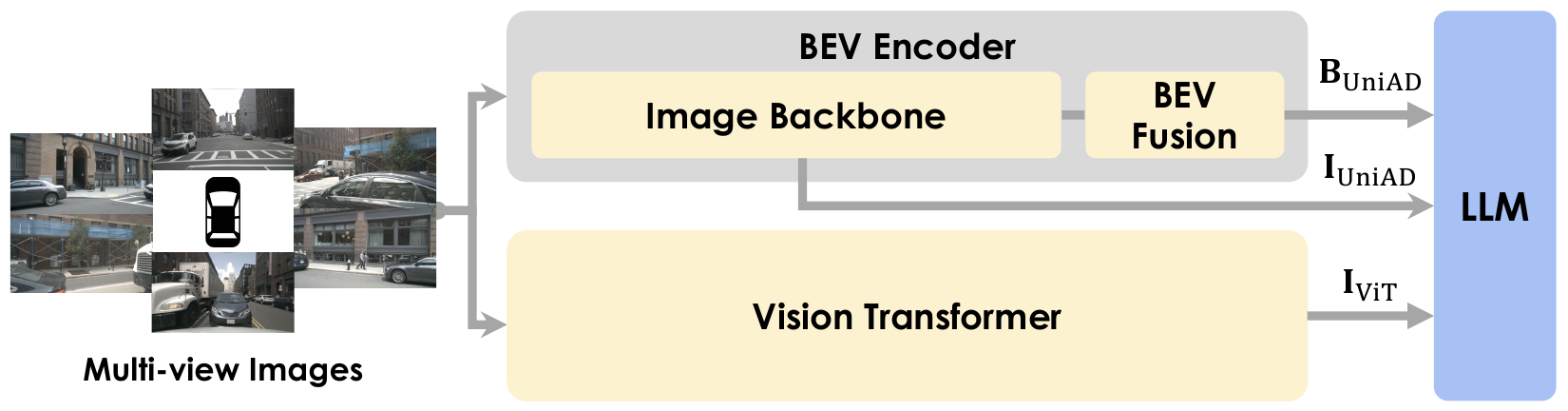}
    \caption{\textbf{Representation Study.} We compare between (1) I$_{\text{ViT}}$, visual tokens extracted from the Vision Transformer of the original VLM; (2) I$_{\text{UniAD}}$, visual tokens from the backbone before the BEV fusion; and (3) B$_{\text{UniAD}}$, BEV tokens produced from the same backbone after the BEV fusion. The input language question is the same, but not visualized here for simplicity.}
    \label{fig:representation_study}
\end{figure}

\subsection{Comparative Study of Visual Representations}
\label{sec:comparative-study}

\colorblock{gray!18}{\textit{\textbf{Research Question 2}: Are BEV representations more effective than perspective-view inputs for enabling LLMs to perform spatial reasoning?}}

Building on the successful BEV-to-language alignment, we next compare different visual representations to examine which one better supports spatial reasoning. Specifically, we evaluate three setups, shown in \cref{fig:representation_study}: (1) I$_{\text{ViT}}$, visual tokens from the ViT of the original VLM~\cite{zhu2025internvl3}; (2) I$_{\text{UniAD}}$, tokens from UniAD’s image backbone before BEV fusion; and (3) B$_{\text{UniAD}}$, BEV tokens produced after fusing multi-view features into BEV space~\cite{hu2023planning}. This analysis allows us to isolate the advantages of the BEV representation over conventional image-based inputs in capturing geometric and spatial relationships within driving scenes. More implementation details can be found in Appendix~\ref{supp:more_implementation_details}.

\subsubsection{Single-View Reasoning.}
We start with DriveLM \textit{object-existence} questions, which are designed for single-view reasoning. The quantitative results are presented in \cref{tab:blm_vqa}. We first observe that BEV representations show advantages over single-view inputs. Additionally, increasing the size of the LLM from 1B to 8B gives a substantial improvement from \SI{89.8}{\percent} to \SI{94.5}{\percent} accuracy, showing that a more powerful LLM can better utilize relevant information from the token space. The 1B LLM processing tokens from B$_{\text{UniAD}}$ achieves \SI{90.8}{\percent} accuracy, which outperforms both I$_{\text{ViT}}$ (\SI{90.3}{\percent}) and I$_{\text{UniAD}}$ (\SI{89.8}{\percent}). This confirms the hypothesized benefits of the BEV representation. Also here, using the larger 8B LLM yields a substantial improvement to \SI{95.3}{\percent} accuracy, outperforming the equivalent model on I$_{\text{UniAD}}$.  Similarly, DeepSeek-VL$_{\text{1B}}$ shows the same trend where the B$_{\text{UniAD}}$ surpasses the I$_{\text{ViT}}$ and I$_{\text{UniAD}}$.

\begin{table}[t]
  \centering
  \caption{\textbf{DriveLM Results and Generality Analysis}. We report accuracy, precision, recall, and F-1 scores. ``\textbf{I}'' denotes image-space, while ``\textbf{B}'' denotes BEV-space. ``Proj.'' indicates training the MLP projector. Metrics are reported as percentages.}
  \vspace{-0.3cm}
  \label{tab:side_by_side_vqa}
  \begin{subtable}{0.48\linewidth}
    \centering
    \caption{InternVL3 Results}
    \label{tab:blm_vqa}
    \resizebox{\linewidth}{!}{
    \begin{tabular}{l|c|c|cccc}
      \toprule
      \textbf{Models} & Mod. & Proj. & \textbf{Acc.}$\uparrow$ & \textbf{Prec.}$\uparrow$ & \textbf{Recall.}$\uparrow$ & \textbf{F1}$\uparrow$ \\
      \midrule
      InternVL3$_\text{1B}$ & I$_\text{ViT}$ &  & 74.2 & 72.2 & 97.7 & 83.0 \\
      InternVL3$_\text{1B}$ & I$_\text{ViT}$ & \checkmark & 90.3 & 90.0 & 95.5 & 92.7 \\
      \midrule
      InternVL3$_\text{1B}$ & I$_\text{UniAD}$ & \checkmark & 89.8 & 90.1 & 94.6 & 92.3 \\
      InternVL3$_\text{8B}$ & I$_\text{UniAD}$ & \checkmark & 94.5 & 94.6 & 97.0 & 95.8 \\
      \midrule
      \rowcolor{bev_blue!13} InternVL3$_\text{1B}$ & B$_\text{UniAD}$ & \checkmark & 90.8 & 90.3 & 96.1 & 93.1 \\
      \rowcolor{bev_yellow!13} InternVL3$_\text{8B}$ & B$_\text{UniAD}$ & \checkmark & 95.3 & 95.2 & 97.7 & 96.4 \\
      \bottomrule
      \multicolumn{7}{c}{ } \\
    \end{tabular}}
  \end{subtable}
  \hfill 
  \begin{subtable}{0.48\linewidth}
    \centering
    \caption{DeepSeek-VL Results}
    \label{tab:blm_vqa2}
    \resizebox{\linewidth}{!}{
    \begin{tabular}{l|c|c|cccc}
      \toprule
      \textbf{Models} & Mod. & Proj. & \textbf{Acc.}$\uparrow$ & \textbf{Prec.}$\uparrow$ & \textbf{Recall.}$\uparrow$ & \textbf{F1}$\uparrow$ \\
      \midrule
      DS-VL$_\text{1B}$ & I$_\text{ViT}$ &  & 61.5 & 84.1 & 49.8 & 62.6 \\
      DS-VL$_\text{1B}$ & I$_\text{ViT}$ & \checkmark  & 85.3 & 86.5 & 91.5 & 89.0 \\ 
      \midrule
      DS-VL$_\text{1B}$ & I$_\text{UniAD}$ & \checkmark  & 90.4 & 89.9 & 97.3 & 93.5 \\
      \midrule
      \rowcolor{bev_green!50} DS-VL$_\text{1B}$ & B$_\text{UniAD}$ & \checkmark & 92.2 & 90.5 & 98.2 & 94.2 \\ 
      \bottomrule
      \multicolumn{7}{c}{ } \\
      \multicolumn{7}{c}{ } \\
      \multicolumn{7}{c}{ } \\
    \end{tabular}}
  \end{subtable}
  \vspace{-0.3cm}
\end{table}

Therefore, even in the context of single-view reasoning tasks, our empirical findings indicate that the BEV representation exhibits a consistent advantage over perspective-view counterparts. This observation motivates a further investigation into the representation gap when subjected to the demands of complex, cross-view spatial reasoning.

\begin{table}[t]
 \centering
 \caption{\textbf{Ego3D Results.} We evaluate cross-view reasoning question types (i.e., ``\textit{object-centric}'' in Ego3D) and report accuracy for Multiple-Choice-Question (MCQ) and L1 error for numerical distance estimation. We use the InternVL3$_\text{1B}$ as the LLM. ``\textbf{Enc. Size}'' and ``\textbf{\#Token}'' denote the visual encoder size and the number of visual tokens, respectively. Token number details can be found in Appendix~\ref{supp:representation-study}.}
 \label{tab:ego3d}
\resizebox{0.75\linewidth}{!}{
    \begin{tabular}{l|cc|ccc|c|c}
    \toprule
    & & & \multicolumn{4}{c|}{\textbf{MCQ Accuracy (\%) $\uparrow$}} & \textbf{L1 (m) $\downarrow$} \\
    \cmidrule(lr){4-7} \cmidrule(lr){8-8}
    \textbf{Modality} & \textbf{Enc. Size} & \textbf{\#Token} & Loc. & Abs-Dist. & Rel-Dist. & \textbf{Avg.} & Dist \\
    \midrule
    I$_{\text{ViT}}$        & 400M & 4,608 & 13.95 & 23.81 & 52.94 & 28.57 & 14.15 \\
    I$_{\text{ViT}}$ w/ ft.& 400M & 4,608 & \underline{60.47} & \textbf{50.00} & \textbf{79.41} & \textbf{62.19} &  \underline{7.42} \\
    \midrule
    I$_{\text{UniAD}}$     & 40M & 2,250 & 39.53 & 26.19 & 64.71 & 42.02 &  9.01 \\
    \rowcolor{bev_blue!13} 
    B$_{\text{UniAD}}$     & 44M & 2,500 & \textbf{67.44} & \underline{45.24} & \underline{73.53} & \underline{61.34} &  \textbf{7.05} \\
    \bottomrule
    \end{tabular}
}
\vspace{-0.3cm}
\end{table}

\subsubsection{Cross-View Reasoning.}
The advantage margin of BEV representation is limited in the DriveLM dataset mainly due to the single-view reasoning questions. Therefore, we show the results on the cross-view reasoning Ego3D dataset~\cite{gholami2025spatial}. Specifically, we focus on the \textit{object-centric} questions, which ask about the spatial relationship between two objects in different ego camera views. Therefore, the LLMs are required to reason over the entire driving scene beyond a single view as in DriveLM perception data. As shown in \cref{tab:ego3d}, we find the BEV representation surpasses the perspective-view representation by a noticeable margin. Specifically, the MCQ accuracy is improved by \SI{46.0}{\percent}, and the L1 error is decreased by \SI{21.8}{\percent}. Additionally, B$_{\text{UniAD}}$ shows comparable performance to I$_{\text{ViT}}$ w/ ft. even though the ViT encoder is 10$\times$ larger. The results further confirm the advantages of BEV representation for complex spatial reasoning. Data examples are provided in Appendix~\ref{sec:ego3d-qa-example}.

One concern is that the BEV fusion introduces extra parameters (e.g., spatial attention~\cite{li2022bevformer}), confounding comparisons between B$_{\text{UniAD}}$ and I$_{\text{UniAD}}$. However, we argue this impact is minor as (1) the module size is only \SI{10}{\percent} of the backbone~\cite{he2016deep}, and (2) B$_{\text{UniAD}}$ shows comparable performance to I$_{\text{ViT}}$ despite the ViT model size being an order of magnitude larger. Thus, the gains mainly stem from BEV’s representational advantages rather than added capacity. Therefore, we can conclude that BEV representation significantly improves the LLM's spatial reasoning capability, especially for panoramic scene understanding.
\begin{figure}[t]
    \centering
    \includegraphics[width=0.75\linewidth]{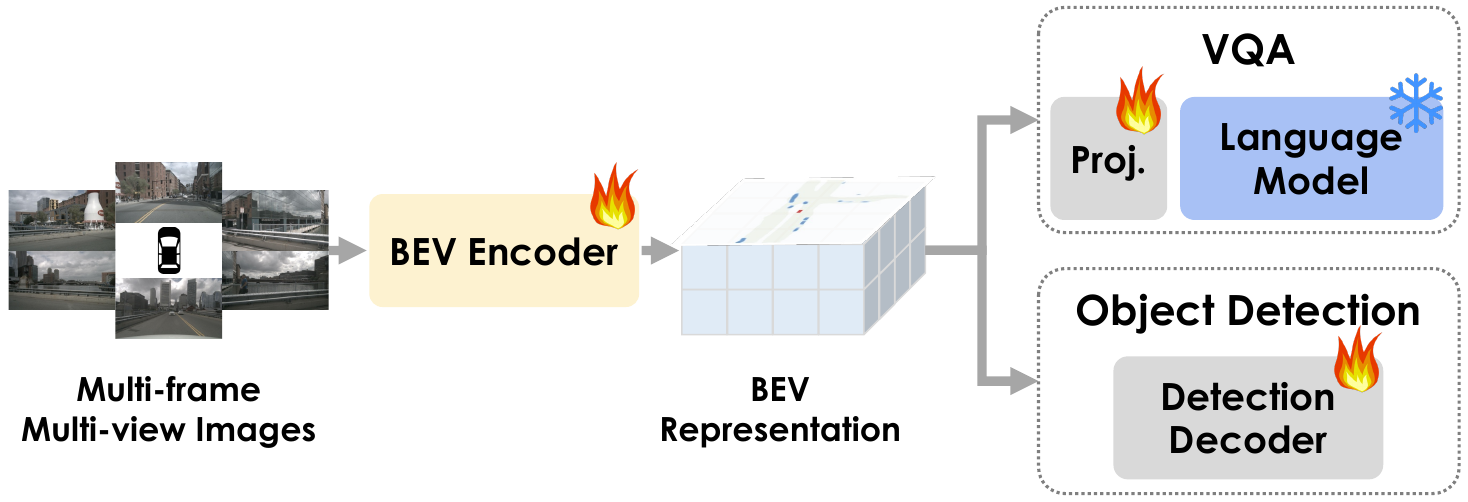}
    \vspace{-0.1cm}
    \caption{\textbf{BEV Semantic Distillation}: We distill the knowledge from the language model to the BEV representations by using Visual Question Answering (VQA) tasks while regularizing BEV spatial structure using the original object detection tasks.}
    \label{fig:bev-distillation}
    \vspace{-0.3cm}
\end{figure}

\section{BEVLM for Semantic Distillation}
\label{sec:approach}

\colorblock{gray!18}{\textit{\textbf{Research Question 3}: Can we enhance the semantic awareness of BEV representations, and thereby improve driving safety, by distilling commonsense knowledge from LLMs into BEV encoders?}}

The representation study in~\cref{sec:comparative-study} confirmed that BEV tokens show superior spatial consistency and geometric coherence compared to perspective-view image tokens for enabling better LLM spatial reasoning. However, BEV encoders are typically trained solely on dense geometric annotations (\eg, bounding boxes~\cite{li2022bevformer}, map elements~\cite{liao2023maptr}, or ego trajectories~\cite{hu2023planning}), ignoring the broader semantic knowledge and commonsense reasoning essential for safety-critical corner cases. This is the key bottleneck that prevents a wide adoption of BEV representations for LLMs, compared to semantically rich 2D visual representations~\cite{sima2023drivelm, shao2024lmdrive, tian2024drivevlm, zhou2025autovla}.

Therefore, we pioneer the exploration of distilling semantic knowledge from pre-trained LLMs into the BEV encoder. By training the BEV encoder using the BEVLM framework to perform high-level visual question answering (VQA)~\cite{antol2015vqa} on safety-related topics, the encoder is \textit{forced} to capture comprehensive, safety-related aspects of the scenario that cannot be covered by geometric supervision alone. This novel semantic distillation process yields a semantically-enhanced BEV representation that is universally beneficial for downstream tasks, including improved performance in LLM-based reasoning~\cite{sima2023drivelm} and enhanced planning in end-to-end (E2E) driving systems~\cite{hu2023planning, jiang2023vad}.

In this work, we focus on evaluating the benefits of distilled BEV representations on the E2E driving pipeline. Since directly using LLMs for E2E control (\eg, VLAs) remains limited by real-time efficiency~\cite{hwang2024emma, zhou2025autovla} and reliability~\cite{xie2025drivebench, ding2025edct}, established E2E pipelines~\cite{hu2023planning, jiang2023vad} remain the most practical architectures for planning, and demonstrating improvement on them provides the most critical evidence of our method's real-world value. We leave LLM-based control (\eg, VLAs) as future work.

\subsection{BEV Distilled from LLM Semantic Space}

As illustrated in \cref{fig:bev-distillation}, we use a shared BEV representation to support both the VQA and object detection tasks. Unlike output-level distillation~\cite{hinton2015distilling}, our method performs representation distillation~\cite{heo2019comprehensive}, where the BEV encoder learns to encode semantic cues directly into its feature grid to satisfy the LLM's reasoning requirements. Specifically, we frame the LLM as a fixed semantic teacher that provides supervision signals via VQA tasks. The BEV encoder (student) is distilled to produce features that align with the semantic space learned by the teacher LLMs. For example, when the LLM answers a question such as \textit{``What is the safe action for the ego vehicle to take?''}, the BEV tokens \textit{must} encode safety-relevant scene information. 

To maintain the spatial structure of the BEV representation, we jointly train with the original perception tasks, such as object detection, which preserve the geometric structure of the BEV grid. This prevents catastrophic forgetting of spatial relationships that are not directly constrained by the distillation.

\noindent
\textbf{Coordinate Conversion.}
Since BEV tokens require reasoning in an ego-centric spatial frame, we convert the image-plane object locations used by DriveLM~\cite{sima2023drivelm, xie2025drivebench} into the ego-centric BEV frame. For instance, instead of expressing a location as \textit{``Object appears at (\SI{450}{px}, \SI{360}{px})''}, we reformulate it as \textit{``Object appears 3 meters in front of the ego vehicle and 1.5 meters to the left''}, enabling more intuitive spatial reasoning. We obtain this by projecting the 3D ground-truth boxes onto the image plane and matching them via Intersection-over-Union, retaining only matches above a predefined threshold.

\noindent
\textbf{Intuitive Interpretation: BEV as a Semantic Manifold.}
Unlike standard auxiliary multi-task supervision, where a task-specific head is learned from scratch, we freeze the LLM parameters $\phi$ so that it acts as a fixed semantic teacher rather than a trainable decoder. The BEV encoder (student) must therefore reshape its feature space to satisfy the LLM's requirements. Concretely, let $\mathcal{X}$ be the input driving scene and $\mathbf{B}_s = E_\theta(\mathcal{X})$ the student BEV feature. For a safety-critical query $\mathbf{q}$, the frozen LLM implicitly requires a specific set of ideal semantic token embeddings $\mathbf{v}^*$, encoding concepts such as ``blocked lane'' or ``unsafe velocity'', to produce the correct answer $\mathbf{a}$. The distillation objective forces the student to align its projected features with $\mathbf{v}^*$:
\begin{equation}
    \mathcal{L}_{\mathrm{distill}} \approx \| \mathrm{MLP}(E_\theta(\mathcal{X})) - \mathbf{v}^* \|_2^2.
\end{equation}

Since $\mathbf{v}^*$ cannot be accessed directly, we use the frozen LLM's cross-entropy loss as a differentiable proxy. Therefore, the VQA dataset is not the end goal, but it acts as an \textit{information bottleneck}: by restricting supervision to complex, reasoning-heavy queries, we selectively distill only the high-level semantics that are absent from the BEV encoder's geometric training. Thus, both the teacher LLM (studied in \cref{sec:close-loop-eval}) and the VQA data (studied in \cref{sec:vqa-ablation}) play a critical role in the proposed framework.

\noindent
\subsection{Training}
Following our previous setup, we use DriveLM-nuScenes~\cite{sima2023drivelm} training split as the VQA data for the distillation. We leverage the \textit{full} VQA data, including perception, prediction, behavior, and planning. We also conduct an ablation study for different VQA data types in \cref{sec:vqa-ablation}. We start from the pre-trained BEVFormer checkpoints~\cite{li2022bevformer} because UniAD~\cite{hu2023planning} and VAD~\cite{jiang2023vad} share the same BEV encoder as BEVFormer and are initialized with the same weights. For the baseline model, we freeze the pre-trained BEV encoder and only train the task-specific decoder. To demonstrate the benefits of our distilled BEV for final E2E driving performance, we introduce this semantic distillation as an additional step after object detection pretraining. The distillation is performed for 1 epoch with equal loss term weights for distillation and object detection. After distillation, we perform the same E2E training as in~\cite{hu2023planning, jiang2023vad}, freezing the distilled BEV encoder and training all task-specific heads for 20 epochs (UniAD) or 12 epochs (VAD). Thus, we can study the effects of semantic enhancement on the final E2E tasks.
All the experiments are performed on 8 NVIDIA A100 80GB GPUs. Distillation for 1 epoch of the full DriveLM dataset takes $\sim$\SI{35}{\hour} for the 1B and \SI{100}{\hour} for the 8B LLM. E2E training time of the UniAD remains unchanged with \SI{115}{\hour}.
\section{Experiments}
\label{sec:experiment}

\subsection{Experimental Setups}

Following prior works~\cite{hu2023planning, jiang2023vad}, we first evaluate the L2 error on the open-loop nuScenes validation set, and further evaluate on the closed-loop NeuroNCAP benchmark~\cite{ljungbergh2024neuroncap}. To account for the stochasticity of the closed-loop simulation, all NeuroNCAP results (score and collision rate) are reported as mean$\pm$std over 50 random-seed runs of the NeuroNCAP scenarios.

\begin{table}[t]
  \centering
  \caption{\textbf{Open-Loop and Closed-Loop E2E Planning Results.} We compare UniAD and VAD performance using the baseline encoder versus the distilled BEV encoder. Open-loop L2 error is measured on nuScenes \cite{caesar2020nuscenes}, while NeuroNCAP score and Collision Rate (CR) are measured on the NeuroNCAP benchmark \cite{ljungbergh2024neuroncap}.}
  \label{tab:e2e_merged_unified}
  \resizebox{0.9\linewidth}{!}{
\begin{tabular}{l|l|ccc|c|cc}
  \toprule
  & & \multicolumn{4}{c|}{\textbf{Open-Loop (nuScenes)}} & \multicolumn{2}{c}{\textbf{NeuroNCAP}} \\
  \cmidrule(lr){3-6} \cmidrule(lr){7-8}
  \textbf{Pipeline} & \textbf{Method} & L2@1s$\downarrow$ & L2@2s$\downarrow$ & L2@3s$\downarrow$ & \textbf{Avg.L2}$\downarrow$ & \textbf{Score}$\uparrow$ & \textbf{CR}$\downarrow$ \\
  \midrule
  \multirow{6}{*}{UniAD}
  & Baseline BEV
  & 0.50 & 0.99 & 1.67 & 1.05 & 2.38$\pm$1.52 & 0.56$\pm$0.31 \\
  & Rand.\ aux.\ head
  & 0.59 & 1.09 & 1.79 & 1.16 & 1.75$\pm$0.96 & 0.73$\pm$0.19 \\
  & VLM-AD trans.
  & 0.56 & 1.02 & 1.64 & 1.07 & 2.02$\pm$1.18 & 0.70$\pm$0.23 \\
  & VLM-AD CLIP
  & 0.57 & 1.01 & 1.64 & 1.07 & 2.07$\pm$1.13 & 0.67$\pm$0.23 \\
  & \cellcolor{bev_blue!13}Distilled BEV$_{\text{1B}}$
  & \cellcolor{bev_blue!13}\textbf{0.46} & \cellcolor{bev_blue!13}\textbf{0.91} & \cellcolor{bev_blue!13}\textbf{1.55} & \cellcolor{bev_blue!13}\textbf{0.97}
  & \cellcolor{bev_blue!13}2.93$\pm$0.69 & \cellcolor{bev_blue!13}0.54$\pm$0.19 \\
  & \cellcolor{bev_yellow!13}Distilled BEV$_{\text{8B}}$
  & \cellcolor{bev_yellow!13}0.48 & \cellcolor{bev_yellow!13}0.94 & \cellcolor{bev_yellow!13}1.59 & \cellcolor{bev_yellow!13}1.00
  & \cellcolor{bev_yellow!13}\textbf{3.05$\pm$1.26} & \cellcolor{bev_yellow!13}\textbf{0.47$\pm$0.30} \\
  \midrule
  & Baseline BEV  
  & 0.47 & 0.79 & 1.19 & 0.82 & 3.24$\pm$1.21 & 0.37$\pm$0.24 \\
  \multirow{-2}{*}{VAD}  
  & \cellcolor{bev_yellow!13}Distilled BEV$_{\text{8B}}$  
  & \cellcolor{bev_yellow!13}\textbf{0.40} & \cellcolor{bev_yellow!13}\textbf{0.71} & \cellcolor{bev_yellow!13}\textbf{1.11} & \cellcolor{bev_yellow!13}\textbf{0.74}
  & \cellcolor{bev_yellow!13}\textbf{3.42$\pm$0.97} & \cellcolor{bev_yellow!13}\textbf{0.34$\pm$0.20} \\
  \bottomrule
\end{tabular}}
  \vspace{-0.1cm}
\end{table}

\subsection{Open-Loop Evaluation}
The open-loop E2E results are evaluated on the nuScenes validation set~\cite{caesar2020nuscenes}.
The results on the L2 error metric at \SI{1}{\second}, \SI{2}{\second}, and \SI{3}{\second} are shown in \cref{tab:e2e_merged_unified}. 

On UniAD, both distilled models show a clear improvement over the baseline at all time horizons. The same trend holds on VAD, where the distilled encoder reduces the average L2 from 0.82 to 0.74. Together, these results indicate the potential of BEVLM distillation in reducing the error of imitating human trajectories across both pipelines.
However, as shown in prior works~\cite{dauner2023parting, li2024ego, dauner2024navsim, cao2025pseudo}, open-loop evaluation provides limited signal about real-world performance. Additionally, the original nuScenes dataset includes scenarios where the ego is mostly moving straight~\cite{li2024ego, xie2025drivebench}. Therefore, the improvement in closed-loop safety-critical scenarios is not well understood.

\subsection{NeuroNCAP Closed-Loop Evaluation}
\label{sec:close-loop-eval}
To study the closed-loop performance in safety-critical scenarios, we further evaluate the distilled model using the NeuroNCAP benchmark \cite{ljungbergh2024neuroncap}.
The benchmark simulates safety-critical scenarios and calculates a NeuroNCAP score between 0 and 5 based on the impact velocity, with 5 being no collision. The NeuroNCAP score provides a more fine-grained evaluation beyond the collision rate metric.

The results are shown in \cref{tab:e2e_merged_unified}. On UniAD, distilling with the 1B LLM raises the NeuroNCAP score substantially from the baseline 2.38 to 2.93 while leaving the collision rate almost unchanged, so the improved score reflects a reduction in crash severity, which we confirm via the pre-collision velocity in Appendix~\ref{supp:more_qualitative_results}. Scaling the teacher to 8B further lifts the score to 3.05 (a \SI{28.2}{\percent} improvement over the baseline) and lowers the collision rate from 0.56 to 0.47, showing that a stronger teacher transfers richer semantics under the same VQA data. To verify that these gains are not specific to UniAD, we transfer the same 8B-distilled BEV encoder to VAD~\cite{jiang2023vad} by freezing it and retraining only the VAD decoder: the NeuroNCAP score improves from 3.24 to 3.42 and the collision rate drops from 0.37 to 0.34. This confirms that semantic distillation generalizes across BEV-based end-to-end pipelines and consistently improves safety in closed-loop, safety-critical scenarios.

We visualize the results between the baseline model and our distilled version (8B) in \cref{fig:neuroncap_all}. The distilled model consistently demonstrates safer and more adaptive behavior in complex scenarios. 
In the first corner case scenario, the ego vehicle turns right into a lane blocked by an excavator. The baseline model shows issues with understanding the scene and proceeds hesitantly, resulting in a collision with the white vehicle approaching from behind. In contrast, the distilled model anticipates the blockage and performs a swift lane change before the white car approaches. 
In the second corner case scenario, a white car drives on the wrong side of the road, approaching the ego. The baseline model reacts too late and causes a crash while steering into the opposing road. Our distilled model, however, avoids the collision by swiftly changing into the free lane on the right, showing a safety-oriented understanding of the scene.
These results highlight that semantic distillation from LLMs equips the BEV encoder with improved situational understanding and safety awareness beyond simply imitating human trajectories in common scenarios~\cite{li2024ego}.
More qualitative examples on the NeuroNCAP benchmark are given in Appendix~\ref{supp:more_qualitative_results}.

\subsection{Comparison with Alternative Semantic Supervision}
\label{sec:alt-supervision}
We compare BEVLM against two alternative supervision designs that use the same DriveLM VQA data: a multi-task learning baseline and a VLM-AD-style distillation~\cite{xu2024vlm}. Both are trained on the UniAD pipeline and reported in \cref{tab:e2e_merged_unified}.

\subsubsection{Multi-Task Learning Baseline.}
\label{sec:rand-aux-head}
The key question is whether the closed-loop improvement comes from the semantics of the pretrained LLM, or simply from attaching an auxiliary VQA head that supervises the BEV encoder with a multi-task objective. To answer this, we replace the frozen LLM with a randomly initialized Transformer decoder of identical architecture ($\sim$13M effectively trainable parameters, matched to BEVLM's $\sim$13M projector) and train it on the same VQA data (\textit{Rand.\ aux.\ head} in \cref{tab:e2e_merged_unified}). Lacking pretrained semantics, this head merely memorizes answer templates and back-propagates no useful semantic gradient to the BEV encoder. As a result, the NeuroNCAP score \emph{drops} from the baseline 2.38 to 1.75, and the collision rate rises from 0.56 to 0.73. This confirms that the gain originates from the pretrained LLM's semantic latent space, not from the auxiliary VQA head itself.

\subsubsection{VLM-AD-Style Distillation.}
\label{sec:vlmad}
We further compare against VLM-AD~\cite{xu2024vlm}, a representative method that distills semantics into the BEV encoder from text supervision. Since VLM-AD and its supervision data are not publicly available, we follow its original recipe: the BEV features are supervised against a CLIP text embedding of answers auto-labeled by InternVL3$_{\text{8B}}$~\cite{zhu2025internvl3} on DriveLM, using a cosine-similarity loss (\textit{VLM-AD CLIP}). We also consider a setup where we use the same auto-labeled answer with a randomly initialized transformer for the distillation (\textit{VLM-AD trans.}). The results are shown in \cref{tab:e2e_merged_unified}. These two variants of VLM-AD only reach a NeuroNCAP score of 2.02 and 2.07, respectively, below the baseline 2.38 and far below BEVLM's 3.05, improving only the open-loop L2 at the \SI{3}{\second} horizon. We attribute this gap to two factors. First, the distillation is performed in the CLIP latent space, which is less semantically rich than the LLM's latent space and thus transfers weaker driving-relevant semantics to the BEV encoder. Second, aligning BEV features to a fixed CLIP embedding integrates the BEV representation with the language model less seamlessly than BEVLM's token-level distillation, which produces features that the LLM can natively consume.

\subsection{Ablation Studies}
\begin{table*}[t]
 \centering
    \caption{\textbf{Ablation on BEV Token Downsampling.} Comparison of pooling and lightweight convolutional methods for BEV downsampling. We evaluate across different tasks and metrics (all scores in \%). The metric abbreviations are: \textbf{B} = BLEU-4~\cite{papineni2002bleu}, \textbf{R} = ROUGE-L~\cite{lin2004rouge}, \textbf{M} = METEOR~\cite{banerjee2005meteor}, \textbf{C} = CIDEr~\cite{vedantam2015cider}, \textbf{Avg.} = Averaged accuracy. We evaluate ``Yes'' or ``No'' questions using Accuracy, while other open-ended questions use language metrics. \textit{Max.~Pool} achieves comparable accuracy with no added parameters and is used in all experiments.}
 \label{tab:projector_ablation}
 \resizebox{\textwidth}{!}{
 \begin{tabular}{r|c|cccc|ccccc|c}
  \toprule
  \textbf{Models} & Projector &  B$\uparrow$& R$\uparrow$& M$\uparrow$&C$\uparrow$& cars &  peds. & trucks & cones & barriers & \textbf{Avg.}$\uparrow$
  \\
  \midrule
    InternVL3$_\text{1B}$& Convolution & 0.469 & \textbf{0.686} & \textbf{0.610} & 3.747 & 94.4 & 90.0 & 84.6 & 86.9 & 90.4 & 91.0 \\ 
    InternVL3$_\text{1B}$ & Depthw. Conv & 0.461 & 0.680 & 0.604 & 3.657 & 94.9 & 88.2 & 85.5 & 89.4 & 89.3 &  90.6 \\ 
    InternVL3$_\text{1B}$ & Concat & 0.454 & 0.673 & 0.597 & 3.621 & 93.8 & 88.2 & 86.4 & 89.4 & 90.0 & 90.2 \\ 
    InternVL3$_\text{1B}$ & Avg. Pool & 0.472 & \textbf{0.686} & 0.609 & \textbf{3.796} & 94.8 & 89.1 & 85.2 & 89.7 & 90.8 & 90.9 \\ 
    \rowcolor{bev_blue!13} InternVL3$_\text{1B}$ & Max. Pool & 0.468 & 0.682 & 0.606 & 3.765 & 94.7 & 88.9 & 85.8 & 90.1 & 88.6 & 90.8 \\ 
    \midrule
    InternVL3$_\text{8B}$ & Concat & 0.468 & 0.661 & 0.567 & 3.521 & 95.6 & 91.1 & 87.3 & 89.4 & 91.5 & 92.4 \\ 
    InternVL3$_\text{8B}$ & Avg. Pool & 0.441 & 0.627 & 0.559 & 3.646 & 97.2 & \textbf{95.5} & \textbf{91.5} & 93.6 & \textbf{96.4} &  \textbf{95.3} \\ 
    \rowcolor{bev_yellow!13} InternVL3$_\text{8B}$ & Max. Pool & \textbf{0.485} & 0.673 & 0.579 & 3.703 & \textbf{97.7} & 94.8 & 89.7 & \textbf{94.0} & 95.0 & \textbf{95.3} \\ 
  \bottomrule
 \end{tabular}}
\end{table*}
\subsubsection{BEV Token Downsampling Method.}
\label{sec:ablation-downsampling}
We conduct an ablation study to determine the optimal BEV downsampling method for the projector, with results for six variants presented in \cref{tab:projector_ablation}.
Given the objective of a lightweight projector, only simple sampling methods are evaluated, supporting the hypothesis that the BEV tokens are already highly expressive and suitable for direct use.
Parameter-free methods include \textit{Avg.~Pool}, \textit{Max.~Pool}, and a \textit{Concat} operation (equivalent to a pixel unshuffle used in VLMs~\cite{zhu2025internvl3}), which substantially increases the projector's parameter count.
Additionally, several learnable methods for downsampling are assessed: a 1-layer standard \textit{Conv}, and a more parameter-efficient \textit{Depthw.~Conv}~\cite{chollet2017xception}.
Same as before, accuracy is measured for existence-based questions. Additionally, we employ various language metrics for a more precise evaluation of the open-ended questions.
The results in \cref{tab:projector_ablation} demonstrate that while the learnable methods perform well, they offer no significant advantage over simple pooling methods, which are parameter-free and computationally efficient. On the 8B LLM, \textit{Avg.~Pool} and \textit{Max.~Pool} achieve the same accuracy. We selected \textit{Max.~Pool} to create the results in this paper due to slightly better values on the language scores.

\begin{table*}[t]
 \centering
 \caption{\textbf{VQA Data Ablation.} We ablate the performance of the distilled BEV encoder on subsets of the DriveLM dataset. All models have similar collision rates (``CR''), allowing for comparison of the average velocity at impact for collision cases (``Imp. Vel.''). The different variants of BEV distillation significantly reduce impact velocities, resulting in higher NeuroNCAP scores. The \textit{Behavior} and \textit{Planning} questions help more than the \textit{Perception} and \textit{Prediction} questions, and the best result is achieved with all questions combined.}
 \label{tab:e2e_data_ablation}
\resizebox{\textwidth}{!}{
\begin{tabular}{l|cccc|cccc|ccc}
    \toprule
    \textbf{Method} & Perception & Prediction & Behavior & Planning & L2@1s$\downarrow$  & L2@2s$\downarrow$  & L2@3s$\downarrow$ & avg.L2$\downarrow$ & NeuroNCAP Score$\uparrow$ & CR@0.0s$\downarrow$ & Imp. Vel.$\downarrow$ \\
    \midrule
    Baseline BEV  &  &  &  &  & 0.50 & 0.99 & 1.67 & 1.05 & 2.38$\pm$1.52 & 0.56$\pm$0.31 & 6.11 \\ 
    Distilled$_{\text{1B}}$ BEV  & \checkmark & \checkmark  & & & 0.48 & 0.95 & 1.62 & 1.02 & 2.77$\pm$1.11 & 0.54$\pm$0.25 & 4.64 \\ 
    Distilled$_{\text{1B}}$ BEV  &  &  & \checkmark & \checkmark  & 0.48 & 0.94 & 1.62 & 1.01  & 2.75$\pm$1.30 & 0.54$\pm$0.26 & \textbf{3.97} \\ 
    \rowcolor{bev_blue!13} Distilled$_{\text{1B}}$ BEV & \checkmark & \checkmark  & \checkmark & \checkmark & \textbf{0.46} & \textbf{0.91} & \textbf{1.55} & \textbf{0.97} & \textbf{2.93$\pm$0.69} & \textbf{0.54$\pm$0.19} & 4.08 \\ 
    \bottomrule
\end{tabular}
}
\vspace{-0.3cm}
\end{table*} 
\begin{figure*}[t]
    \centering
    \begin{subfigure}[t]{\textwidth}
        \centering
        \includegraphics[width=\textwidth]{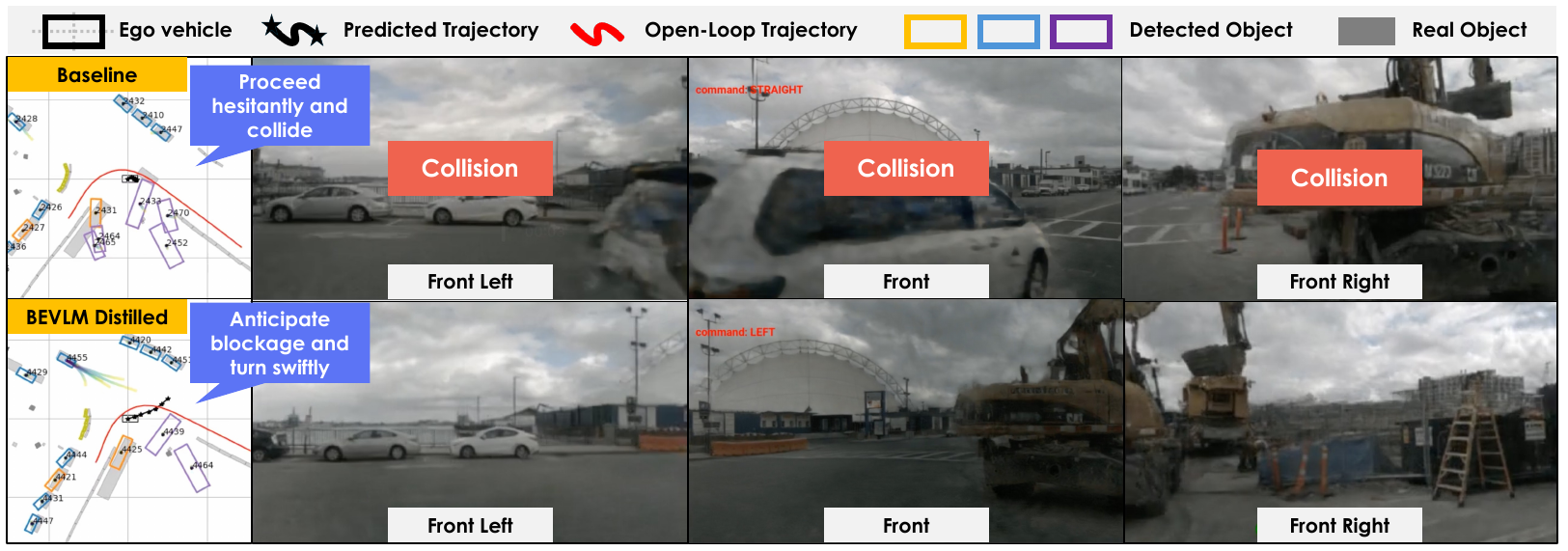}
        \caption{\textbf{Corner Case 1: Right-Turn Conflict with Blocked Lane:} In this scenario, the ego vehicle attempts to turn right at an intersection, but the rightmost lane after the turn is blocked by an excavator. Meanwhile, a vehicle traveling straight in the adjacent lane restricts the available space, requiring the ego to adjust its trajectory to avoid a potential collision.
        }
        \label{fig:neuroncap_1}
    \end{subfigure}
    \hfill
    \begin{subfigure}[t]{\textwidth}
        \centering
        \includegraphics[width=\textwidth]{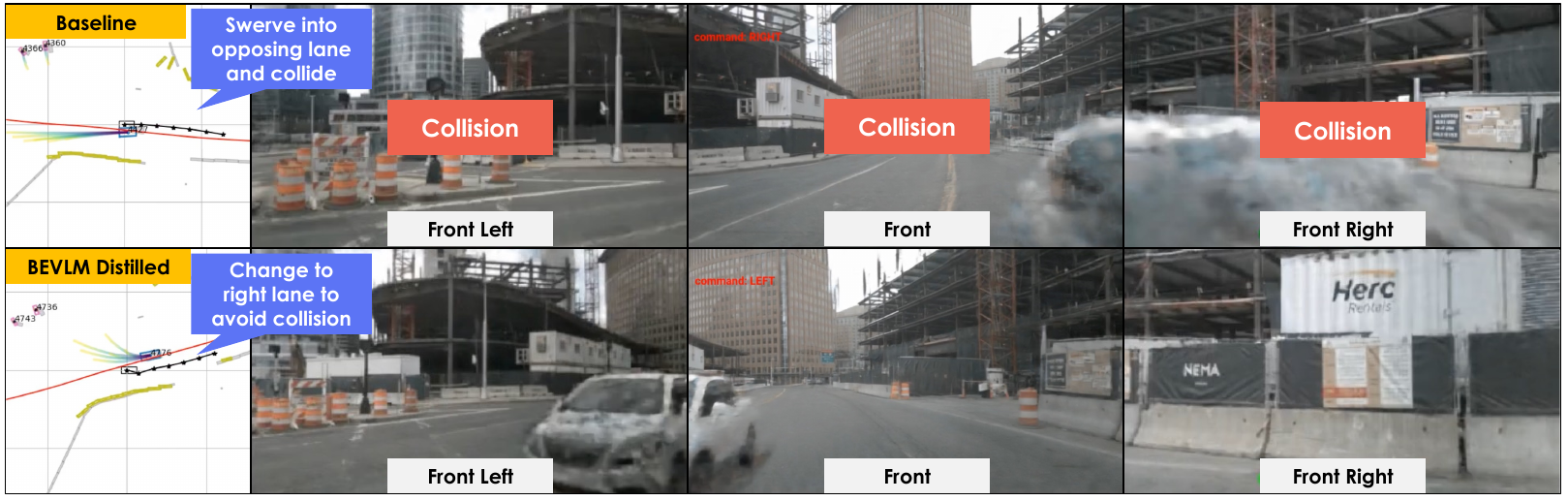}
        \caption{\textbf{Corner Case 2: Oncoming Car in Ego Lane:} In this scenario, the ego vehicle drives on a two-way road with two lanes in each direction. A car from the opposite direction enters the ego vehicle’s lane and drives toward it on the wrong side of the road. To avoid a collision, the ego vehicle needs to move to the free right lane in its own direction.
        }
        \label{fig:neuroncap_2}
    \end{subfigure}
    \caption{\textbf{Qualitative NeuroNCAP Results.} Two representative closed-loop planning scenarios are presented for comparison between the baseline and semantically distilled models. The distilled model demonstrates improved decision-making under safety-critical scenarios, successfully performing a \textit{safe right turn} in corner case 1 and an \textit{evasive lane change to the free right lane} in corner case 2 to avoid potential collisions, where the baseline model fails.}
    \label{fig:neuroncap_all}
\end{figure*}

\subsubsection{Distillation VQA Data.}
\label{sec:vqa-ablation}
We further conduct an ablation study from the VQA data perspective. Specifically, we study what type of question can best benefit the BEV semantic learning by separating the questions in the DriveLM dataset, including \textit{Perception}, \textit{Prediction}, \textit{Behavior}, and \textit{Planning}. Since the different task data samples vary, we train the less frequent subset (\ie, behavior and planning) for 3 epochs while training the more frequent subset (\ie, perception and prediction) for one epoch to match the overall iterations (shown in Appendix~Tab.~\ref{tab:data-stat}).

The results are presented in \cref{tab:e2e_data_ablation}. Both question subsets individually yield a substantial improvement over the baseline with no distillation, reaching comparable NeuroNCAP scores of 2.77 (\textit{Perception} and \textit{Prediction}) and 2.75 (\textit{Behavior} and \textit{Planning}). Combining all four question types achieves the best NeuroNCAP score of 2.93, indicating that the two subsets provide complementary semantic supervision.

The collision rate is similar for all variants (around 0.54), so the higher NeuroNCAP score indicates that the velocity at impact is much lower for the models trained from distilled BEV encoders, indicating safer and more anticipatory behavior in such corner cases.
This is confirmed when comparing the average impact velocity.
The UniAD baseline trained without distillation has \SI{6.11}{\meter\per\second}. Training on the \textit{Perception} and \textit{Prediction} subset reduces the average impact velocity to \SI{4.64}{\meter\per\second}, while training on the \textit{Behavior} and \textit{Planning} subset is even more impactful, reaching the lowest value of \SI{3.97}{\meter\per\second}.
The variant distilled from all DriveLM questions reaches \SI{4.08}{\meter\per\second} while attaining the highest overall NeuroNCAP score, showing that both subsets complement each other.
More qualitative results together with the corresponding velocity profiles are discussed in Appendix~\ref{sec:impact_severity}.
\section{Discussion \& Conclusion}
\label{sec:discussion}

\noindent
\textbf{Limitations.} In this work, we mainly use the DriveLM-nuScenes~\cite{sima2023drivelm} dataset for the experiments. We select this dataset due to its (1) compatibility with nuScenes~\cite{caesar2020nuscenes} and (2) curated and high-quality ground truth. However, this should not be a major concern, as the distillation from this dataset alone already brings significant improvement. Evaluating the distilled BEV representation on more diverse and semantically rich VQA data to confirm the scaling of the framework is left for future work. More discussion is in Appendix~\ref{supp:limitation}.

\noindent
\textbf{Conclusion.} 
In this work, we presented a framework that unifies the spatial structure of BEV representations with the semantic reasoning capabilities of LLMs. We show that BEV features can be effectively tokenized for LLM processing and that they enable stronger spatial reasoning than image-based tokens in multi-view settings. Building on this insight, we introduced \textsc{BEVLM}, which distills semantic knowledge from LLMs into BEV encoders. This semantic enrichment leads to consistent gains in end-to-end driving, including up to \SI{28.2}{\percent} improvement in safety on the NeuroNCAP benchmark. These results highlight the promise of integrating BEV with LLM reasoning to improve the safety and reliability of autonomous driving systems.

\section*{Acknowledgements}

We thank Mukesh Ghimire and Joona Hellmuth for their help with running the NeuroNCAP closed-loop simulation. This work was primarily supported by Mercedes-Benz Research \& Development North America, Inc. Shaoyuan Xie and Qi Alfred Chen were supported in part by (1) the National Science Foundation under grants CNS-2145493 and CNS-2413877, and (2) the U.S. Department of Transportation under Grant 69A3552348327 through the CARMEN+ University Transportation Center.

\newpage

{
    \small
    \bibliographystyle{splncs04}
    \bibliography{main}
}

\title{\texorpdfstring{BEVLM: Distilling Semantic Knowledge from LLMs into Bird's-Eye View Representations\\\vspace{0.2cm} --Supplementary Material--}{BEVLM: Distilling Semantic Knowledge from LLMs into Bird's-Eye View Representations - Supplementary Material}}
\titlerunning{BEVLM}
\authorrunning{T. Monninger, S. Xie, Q. A. Chen, and S. Ding}
\author{}
\institute{}
\edef\bevlmsavefig{\arabic{figure}}
\edef\bevlmsavetab{\arabic{table}}
\edef\bevlmsaveeq{\arabic{equation}}
\maketitle
\setcounter{figure}{\bevlmsavefig}
\setcounter{table}{\bevlmsavetab}
\setcounter{equation}{\bevlmsaveeq}

\renewcommand{\theHsection}{supp.\arabic{section}}
\renewcommand{\theHsubsection}{supp.\arabic{section}.\arabic{subsection}}
\renewcommand{\theHsubsubsection}{supp.\arabic{section}.\arabic{subsection}.\arabic{subsubsection}}

\appendix

\section*{Table of Contents}
\startcontents[appendices]
\printcontents[appendices]{l}{1}{\setcounter{tocdepth}{3}}
\vspace{0.2cm}

\section{More Implementation Details}
\label{supp:more_implementation_details}

\begin{figure}[t]
    \centering
    \includegraphics[width=0.8\linewidth]{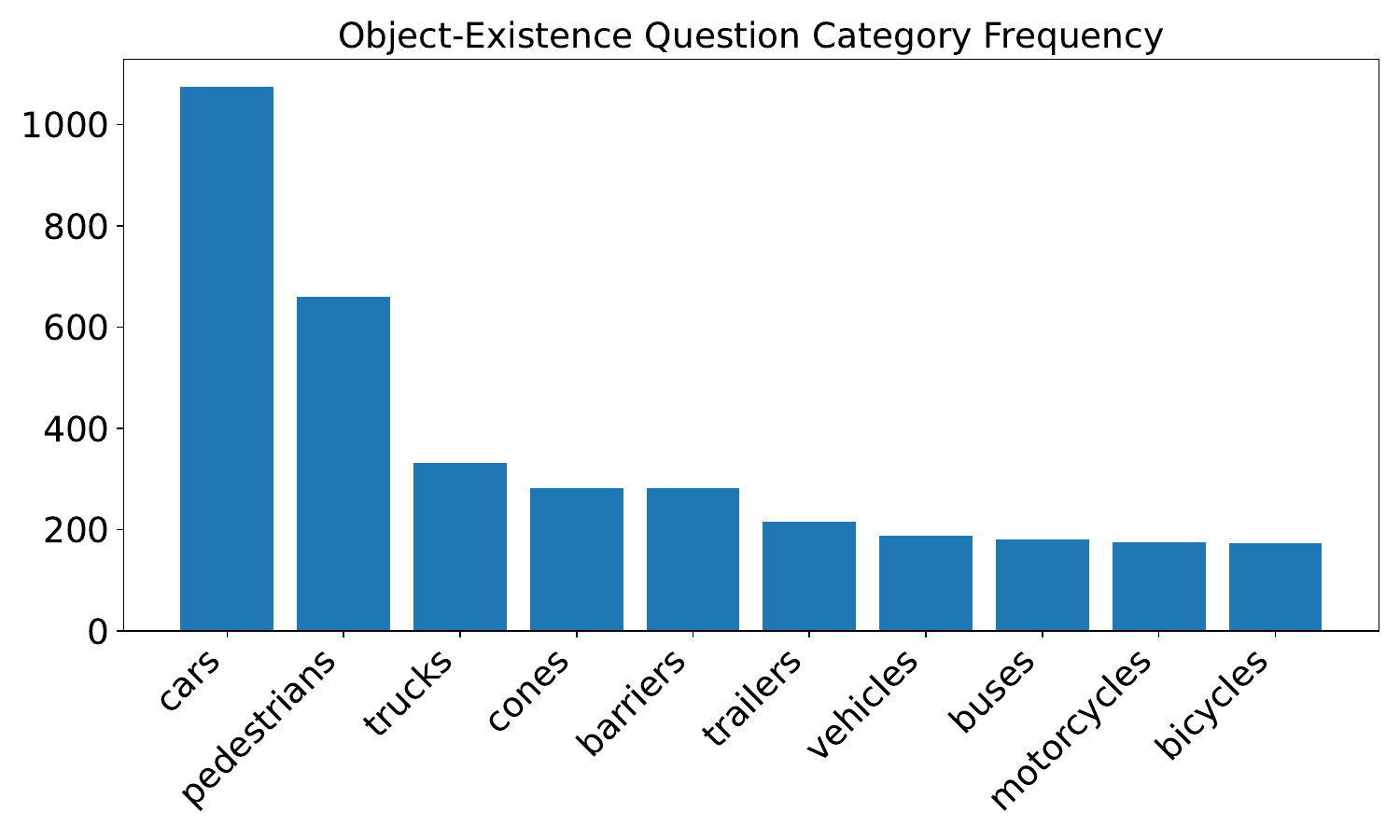}
    \caption{\textbf{DriveLM Dataset Object Frequency.} We visualize the object frequency in the \textit{Perception} object-existence related questions, which are used to compute accuracy in \cref{tab:mlp_projector} and \cref{tab:side_by_side_vqa} in the main paper.}
    \label{fig:obj-frequency}
\end{figure}

In this section, we describe the detailed implementation in the main paper.

\subsection{Dataset Details}
\label{sec:train-detail}

\subsubsection{DriveLM Dataset.}
\label{sec:data-stat}

The DriveLM-nuScenes~\cite{sima2023drivelm} data distribution is shown in \cref{tab:data-stat}. For the results in \cref{tab:mlp_projector} and \cref{tab:side_by_side_vqa} of the main paper, we use only the \textit{Perception} split for training. We use the full dataset for semantic distillation in \cref{sec:approach}.

For the accuracy evaluation questions (\ie, \textit{object-existence} questions), we report the frequency of each object category in \cref{fig:obj-frequency}. We apply filtering for these questions to ensure a fair comparison across methods. Specifically, we observe that some detection-related questions in DriveLM contain incorrect answers or reference objects located far from the ego vehicle. BEV-based models, however, are trained to represent the scene only within a limited spatial extent. For example, BEVFormer~\cite{li2022bevformer} and UniAD~\cite{hu2023planning} use a 3D cuboid centered on the ego vehicle with range $[-51.2, -51.2, -5.0,\, 51.2, 51.2, 3.0]$ along $[-x,-y,-z,\,x,y,z]$ in the ego frame. Therefore, we keep the questions whose referenced ground-truth objects fall within this spatial range. We apply the same filtering to the 3D annotations before determining whether a question can be answered correctly. After filtering, 3{,}609 questions remain for evaluating the \textit{object-existence} accuracy, which is reported in \cref{tab:mlp_projector} and \cref{tab:side_by_side_vqa} of the main paper.

\subsubsection{Ego3D Dataset.} 

The Ego3D dataset~\cite{gholami2025spatial} only includes the evaluation split with 2{,}231 pieces of data originating from the nuScenes dataset~\cite{caesar2020nuscenes}. We split the dataset into training and testing sets with a ratio of 8:2. Specifically, we use 1{,}780 pieces of data as the training set and evaluate on the ``\textit{object-centric}'' question type in the testing set, since it requires cross-view reasoning. The distribution of testing data used in \cref{tab:ego3d} can be found in \cref{tab:data-stat-ego3d}.

\begin{table}[t]
\centering
\caption{\textbf{Detailed training VQA data distribution} of the DriveLM-nuScenes dataset~\cite{sima2023drivelm}. The perception data is mainly used for \cref{tab:mlp_projector} and \cref{tab:side_by_side_vqa} in the main paper, while the full set is used for semantic distillation.}
\vspace{-0.2cm}
\label{tab:data-stat}
\resizebox{0.8\linewidth}{!}{
\begin{tabular}{l|c|c|c}
    \toprule
    \textbf{Driving Task} & \textbf{Question Type} & \textbf{\# Samples} & \textbf{Total}
    \\
    \midrule
    $\circ$ Perception & Yes-or-No \& VQA & 139{,}187 & \multirow{4}{*}{\textbf{298{,}620}}  
    \\
    $\circ$ Prediction & VQA & 85{,}842             
    \\
    $\circ$ Planning & VQA & 69{,}935             
    \\
    $\circ$ Behavior & VQA & 3{,}656
    \\
    \bottomrule
\end{tabular}}
\end{table}

\begin{table}[t]
\centering
\caption{\textbf{Detailed Ego3D testing data distribution}.}
\vspace{-0.2cm}
\label{tab:data-stat-ego3d}
\resizebox{0.8\linewidth}{!}{
\begin{tabular}{l|c|c|c}
    \toprule
    \textbf{Driving Task} & \textbf{Question Type} & \textbf{\# Samples} & \textbf{Total} \\
    \midrule
    $\circ$ Localization & MCQ & 43 & \multirow{4}{*}{\textbf{162}} \\
    $\circ$ Object-Centric Abs. Dist. & MCQ & 42 &  \\
    $\circ$ Object-Centric Rel. Dist. & MCQ & 34 &  \\
    $\circ$ Object-Centric Abs. Dist. & Numerical & 43 &  \\
    \bottomrule
\end{tabular}}
\end{table}

\subsection{Baselines Details}
\label{sec:baseline-detail}

In this section, we describe how the baselines in \cref{tab:mlp_projector} of the main paper are computed. These baselines help evaluate how well BEV features are aligned with the language space.

\subsubsection{Majority Class.}
For this baseline, we examine the training split of the DriveLM dataset. For each \textit{object category} (\eg, car, pedestrian), we collect all questions following the pattern  
\textit{``Is there a \{status\} \underline{\textit{\smash{category}}} in the \{view angle\}?''},  
ignoring the specific \textit{status} and \textit{view angle}. We then count whether ``Yes’’ or ``No’’ appears more often for that category. During testing, for any question referring to the same object category, we always output the majority answer. This serves as a simple prior-based baseline. If a model performs similarly to this baseline, it suggests that the learned projector is mainly capturing dataset bias rather than extracting meaningful information from BEV features, as discussed in prior work~\cite{xie2025drivebench}.

\subsubsection{Linear Probe.}
This section provides more information on the region-aware linear probe designed for the ego-centric spatial questions in DriveLM.
For each question, we generate a binary spatial mask on the BEV plane corresponding to the specified camera view (\eg, ``front-left''), derived from the nuScenes camera extrinsic parameters.
This mask is applied element-wise to the BEV feature grid to isolate features within the relevant field of view.
We employ global max pooling over the spatial dimensions of the masked features, followed by Layer Normalization.
Based on the object class in the question, the resulting vector is then passed to a class-specific linear classifier to predict a binary existence probability. 
This formulation serves as a baseline to isolate the representation's semantic quality from global dataset biases while leveraging location and object information from the question.

\subsubsection{Detection Baseline.}
\label{sec:det-baseline}
We also compare BEVLM against the UniAD detection baseline~\cite{hu2023planning}. This is a task-specific model that uses the same BEV encoder, so both approaches have access to the same underlying spatial information. If our projector preserves most information in BEV features, its performance should approach the detector's performance. Following the standard UniAD setup~\cite{hu2023planning}, we use the detection head to predict 3D bounding boxes. For each test question, we match the question to the predicted boxes using three criteria: \textit{object category}, \textit{moving status}, and \textit{view angle}.

For computing the view angle, we follow the nuScenes camera configuration~\cite{qian2024nuscenes}. The front and side cameras have a \SI{70}{\degree} field of view (FOV), while the rear camera has a \SI{110}{\degree} FOV. Therefore, we classify any object within the ego angle range [\SI{-35}{\degree}, \SI{35}{\degree}] as being in the front view.

To determine the moving status, we follow the nuScenes definition and classify an object as \textit{``moving''} if its absolute predicted speed exceeds \SI{0.2}{\meter\per\second}.

\subsection{Visual Tokenization Implementation}
\label{supp:representation-study}
For the I$_{\text{UniAD}}$ comparison, we extract visual tokens from the lowest-resolution feature maps of the Feature Pyramid Network (FPN)~\cite{lin2017feature}, using the same MLP projector as for BEV features. FPN produces four spatial resolutions: $116\times 200$, $58\times 100$, $29\times 50$, and $15\times 25$, corresponding to the ResNet~\cite{he2016deep} stages in the top-down pathway with skip connections. We select the lowest-resolution map because its spatial size is closest to the BEV token count (about 2{,}500), which makes the comparison between image tokens and BEV tokens more balanced. With six camera views, this results in $15\times 25\times 6 = 2{,}250$ visual tokens.

For the ViT-based comparison, we follow the InternVL3 preprocessing pipeline \cite{zhu2025internvl3}. Each original $900\times 1600$ image is downsampled to $450\times 800$, then split into two $448\times 448$ tiles by rounding the spatial size to the nearest multiple. We additionally enable \texttt{use\_thumbnail=True}, which creates a third tile by resizing the full $450\times 800$ image into a single $448\times 448$ input. Each tile produces 256 visual tokens, giving $256\times 3 = 768$ tokens per camera view. With six views, the ViT setting uses $768\times 6 = 4{,}608$ visual tokens in total.

For DeepSeek-VL~\cite{lu2024deepseekvl}, we use the official implementation. Each image is processed by hybrid visual encoders, including SigLIP~\cite{zhai2023sigmoid} for extracting high-level semantic features and SAM-B~\cite{kirillov2023segment} for low-level cues. As a result, each image is represented by a fixed set of 576 visual tokens, independent of the input resolution. With six views, this yields $576\times 6 = 3{,}456$ tokens per scene.

\subsection{Implementation of BEVLM Framework}
\label{supp:bevlm-detail}

\subsubsection{Distillation Mechanism and Setup.}
To align BEV features with the language space, we follow simple yet effective visual alignment strategies used in VLMs~\cite{liu2023visual, zhu2023minigpt}. Specifically, we employ a lightweight MLP projector to map pre-trained BEV features~\cite{hu2023planning} into BEV tokens. 
We avoid more complex architectures, such as Q-Former~\cite{li2022blip}, to maintain simplicity and ensure a fair comparison with existing MLP-projector VLMs~\cite{zhu2025internvl3, qwen2.5}.

For tokenization, the BEV feature grid $\mathbf{B}\in\mathbb{R}^{H_{\text{BEV}}\times W_{\text{BEV}}\times C}$ can be viewed as a top-down image. Although each grid cell could serve as a BEV token, the native resolution (\eg, $200\times200$) yields extensive tokens. To balance spatial detail and efficiency, we downsample the grid into a compact representation. We introduce two special tokens, \texttt{<bev>} and \texttt{</bev>}, to mark the BEV token sequence. During training, only the projector and these tokens are learned, while the BEV encoder and LLM remain frozen. We use standard next-token prediction conditioned on BEV tokens as the training objective:
\begin{equation}
\mathcal{L}(\theta)
    = - \mathbb{E}_{(\mathbf{q},\, \mathbf{a}) \sim \mathcal{D}}
        \Bigg[\sum_{t=1}^{T}
        \log p_{\theta}\big(\mathbf{a}_t \mid \mathbf{B},\, \mathbf{q}, \mathbf{a}_{1:t-1}\big)
    \Bigg] \nonumber,
\end{equation}
\noindent
where $\theta$ denotes the model parameters, and $\mathbf{q}$ and $\mathbf{a}$ are the question and ground-truth answer.

\subsection{Training Configurations}

\subsubsection{Training Parameters.}

For the experiments in \cref{tab:mlp_projector} and \cref{tab:side_by_side_vqa} of the main paper, we train the projector for 4 epochs to ensure stable convergence. We use a learning rate of $1\times10^{-3}$ with cosine learning-rate decay and the AdamW optimizer~\cite{loshchilov2017decoupled}. For encoder fine-tuning, we start from the trained projector and train for one additional epoch, using a learning rate of $1\times10^{-5}$ for the projector and $1\times10^{-6}$ for the encoder.

\subsubsection{MLP Projector Design.}
We follow the same MLP projector design as that in the InternVL3~\cite{zhu2025internvl3}. Specifically, the MLP projector is composed of \{\texttt{LayerNorm}, \texttt{Linear}, \texttt{GELU}, \texttt{Linear}\}, sequentially. We use the same MLP projector for BEV alignment, both for InternVL3 and DeepSeek-VL~\cite{lu2024deepseekvl}.

\subsubsection{Data Pipeline.} 

We iterate over the question-answer samples while retrieving the corresponding images using the scene token defined in nuScenes~\cite{caesar2020nuscenes}. Therefore, the same set of multi-view images might be used multiple times, depending on the number of questions for that specific scene. 

\begin{table*}[t]
 \centering
 \caption{\textbf{BEV Projector Alignment Study.} We compare the performance of three baselines with the performance of an LLM that operates directly on the BEV grid. We report accuracy individually for each class as an extension of \Cref{tab:mlp_projector}. \textit{Majority class} denotes using the majority choices from the training set for each category (\cref{sec:det-baseline}).}
 \vspace{-0.25cm}
 \label{tab:mlp_projector_full}
 \resizebox{\linewidth}{!}{
 \begin{tabular}{l|c|cccccccccc|c}
  \toprule
  \textbf{Models} & Modality & cars &  pedestrians & trucks & cones & barriers & trailers & const. vehicles & buses & motorcycles & bicycles & \textbf{Average}$\uparrow$
  \\
  \midrule
    Majority class & - & 92.7 & 81.9 & 65.0 & 59.6 & 54.8 & 66.7 & 83.4 & 75.0 & 79.9 & 77.5 & 78.2 \\ 
    Linear probe & - & 94.6 & 87.2 & 84.2 & 83.8 & 83.9 & 95.4 & 87.2 & 85.3 & 84.2 & 84.4 & 88.7 \\
    Detection$_\text{UniAD}$ & - & 91.1 & 90.9 & 86.7 & 99.0 & 99.6 & 92.6 & 95.7 & 96.1 & 94.8 & 93.6 & 92.8 \\
    \midrule
    \rowcolor{bev_blue!13} InternVL3$_\text{1B}$ & B$_\text{UniAD}$ & 94.7 & 88.9 & 85.8 & 90.1 & 88.6 & 94.9 & 90.4 & 90.0 & 90.2 & 85.6 & 90.8 \\
    \rowcolor{bev_yellow!13} InternVL3$_\text{8B}$ & B$_\text{UniAD}$ & 97.7 & 94.8 & 89.7 & 94.0 & 95.0 & 97.2 & 92.0 & 96.7 & 96.0 & 94.8 & 95.3 \\
  \bottomrule
  \end{tabular}}
\end{table*}

\begin{table*}[t]
 \centering
 \caption{\textbf{Scene Representation Results}. We report accuracy individually for each class as an extension of \Cref{tab:side_by_side_vqa}. ``\textbf{I}'' denotes image-space representations, whereas ``\textbf{B}'' denotes BEV-space representations. ``Proj.'' indicates training only the MLP projector, while ``Enc.'' refers to training the modality encoder.
 }
 \vspace{-0.2cm}
 \label{tab:blm_vqa_full}
 \resizebox{\textwidth}{!}{
\begin{tabular}{l|c|cc|cccccccccc|c}
  \toprule
  \textbf{Models} & Modality & Proj. & Enc.  &    cars &  pedestrians & trucks & cones & barriers & trailers & const. vehicles & buses & motorcycles & bicycles&\textbf{Average}$\uparrow$ \\
  \midrule
    InternVL3$_\text{1B}$ & I$_\text{ViT}$ &   &    & 92.8 & 83.5 & 69.8 & 81.6 & 55.2 & 55.1 & 31.0 & 52.2 & 55.8 & 62.4 &74.2 \\ 
    InternVL3$_\text{1B}$ & I$_\text{ViT}$ & \checkmark &    & 95.7 & 88.6 & 84.3 & 89.7 & 88.3 & 89.4 & 89.8 & 85.0 & 87.4 & 87.9 &90.3 \\ 
    InternVL3$_\text{1B}$ & I$_\text{ViT}$ & \checkmark & \checkmark   & 95.5 & 88.8 & 85.2 & 89.4 & 89.0 & 89.8 & 89.8 & 87.8 & 87.4 & 87.9 &90.5 \\ 
    \midrule
    InternVL3$_\text{1B}$ & I$_\text{UniAD}$ & \checkmark &    & 94.8 & 89.4 & 82.2 & 89.4 & 89.0 & 94.9 & 89.8 & 83.3 & 82.2 & 84.4 &89.8 \\
    InternVL3$_\text{1B}$ & I$_\text{UniAD}$ & \checkmark & \checkmark   & 94.3 & 89.2 & 83.7 & 89.7 & 89.7 & 95.4 & 88.8 & 85.6 & 81.6 & 85.6 &89.9 \\
    \midrule
    InternVL3$_\text{8B}$ & I$_\text{UniAD}$ & \checkmark &    & 97.2 & 93.3 & 90.3 & 93.6 & 94.0 & 97.7 & 91.4 & 95.0 & 93.7 & 91.9 &94.5 \\
    InternVL3$_\text{8B}$ & I$_\text{UniAD}$ & \checkmark & \checkmark   & 97.3 & 93.0 & 89.1 & 91.5 & 94.0 & 97.7 & 94.7 & 94.4 & 94.3 & 92.5 &94.4 \\
    \midrule
    \rowcolor{bev_blue!13} InternVL3$_\text{1B}$ & B$_\text{UniAD}$ & \checkmark &   & 94.7 & 88.9 & 85.8 & 90.1 & 88.6 & 94.9 & 90.4 & 90.0 & 90.2 & 85.6 &90.8 \\
    InternVL3$_\text{1B}$ & B$_\text{UniAD}$ & \checkmark & \checkmark & 94.6 & 89.2 & 87.0 & 89.4 & 90.8 & 94.4 & 89.3 & 89.4 & 91.4 & 86.7 &91.1 \\ 
    \midrule
    \rowcolor{bev_yellow!13} InternVL3$_\text{8B}$ & B$_\text{UniAD}$ & \checkmark &  & 97.7 & 94.8 & 89.7 & 94.0 & 95.0 & 97.2 & 92.0 & 96.7 & 96.0 & 94.8 &95.3 \\ 
    InternVL3$_\text{8B}$ & B$_\text{UniAD}$ & \checkmark & \checkmark   & 96.8 & 95.3 & 90.3 & 93.3 & 94.7 & 96.3 & 93.6 & 97.8 & 95.4 & 94.8 &95.2 \\
  \bottomrule
 \end{tabular}
}
\end{table*}

\section{More Experimental Results}
\label{supp:more_experiment_results}

We provide the full class accuracy beyond the most frequent five classes reported in the main paper (\cref{tab:mlp_projector}). The results are illustrated in \cref{tab:mlp_projector_full,tab:blm_vqa_full}. The results show a similar trend that the BEV representation shows consistently better performance compared to the perspective-view representations, which further justifies our claims.
\section{Example Text Data}
\label{supp:example_text_data}

\subsection{DriveLM Example Question-Answer Pairs}
\label{sec:qa-example}

We provide examples of the \textit{object-existence} questions, which are a subset of \textit{Perception} questions that we use to compute the accuracy, as shown below:
\vspace{0.3cm}

\qaside{Are there moving cars to the back right of the ego car?}{Yes.}
\qaside{Are there barriers to the front of the ego car?}{No.}
\qaside{Are there traffic cones to the front left of the ego car?}{No.}
\qaside{Are there parked cars to the back left of the ego car?}{Yes.}
\qaside{Are there bicycles without riders to the front right of the ego car?}{No.}
\qaside{Are there standing pedestrians to the front of the ego car?}{Yes.}

\subsection{Ego3D Example Question-Answer Pairs}
\label{sec:ego3d-qa-example}
We provide examples of the testing questions in Ego3D, which require complex spatial reasoning. The four data points here represent ``\textit{Localization}'', ``\textit{Object-Centric Abs. Dist (MCQ)}'', ``\textit{Object-Centric Rel. Dist. (MCQ)}'', and ``\textit{Object-Centric Abs. Dist. (Numerical)}'' shown in \cref{tab:data-stat-ego3d}.

\qaside{
The front view corresponds to the north direction. Assume the black BMW hatchback stopped at the barrier in the front view is facing north. From the perspective of the black BMW hatchback, where is the ego car located?
    \begin{itemize}
        \item A. Front
        \item B. Back
        \item C. Right
        \item D. Left
    \end{itemize}
}{B}

\qaside{
How far, in meters, is the black BMW hatchback stopped at the barrier in the front view from the pedestrian wearing red and crossing the driveway in the front view?
    \begin{itemize}
        \item A. 36 meters
        \item B. 25 meters
        \item C. 13 meters
        \item D. 3 meters
    \end{itemize}
}{C}

\qaside{
Which one is closer to the black BMW hatchback stopped at the barrier in the front view?
    \begin{itemize}
        \item A. Ego car
        \item B. Pedestrian wearing red crossing the driveway
    \end{itemize}
}{A}

\qaside{
How far, in meters, is the black BMW hatchback stopped at the barrier in the front view from the pedestrian wearing red and crossing the driveway in the front view?
}{13.7 meters}
\section{More Qualitative Results on NeuroNCAP}
\label{supp:more_qualitative_results}

In this section, we provide and analyze further qualitative results from the NeuroNCAP corner case benchmark.

\begin{figure*}[t]
    \centering
    \begin{subfigure}[t]{\textwidth}
        \centering
        \includegraphics[width=\textwidth]{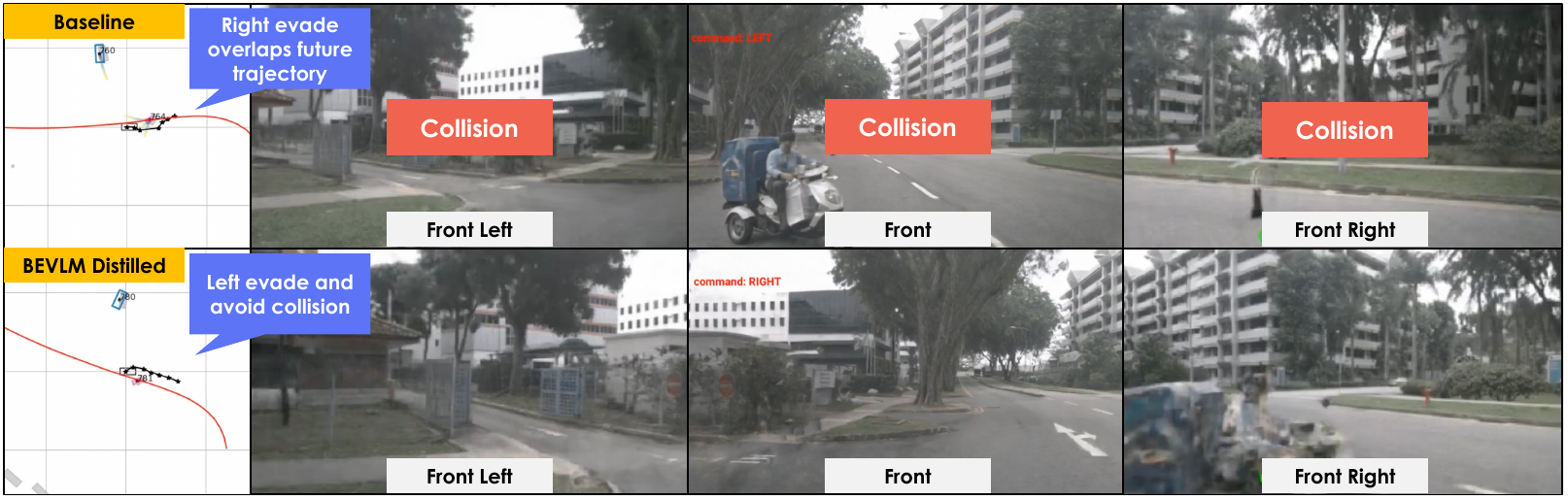}
        \caption{\textbf{Corner Case 3: } In this scenario, the ego vehicle is moving straight while a motorcycle cuts into the lane from the left. To avoid the collision, the ego vehicle is supposed to swerve to the left, avoiding the future moving trajectory of the motorcycle.
        }
        \label{fig:neuroncap_3}
    \end{subfigure}
    \hfill
    \begin{subfigure}[t]{\textwidth}
        \centering
        \includegraphics[width=\textwidth]{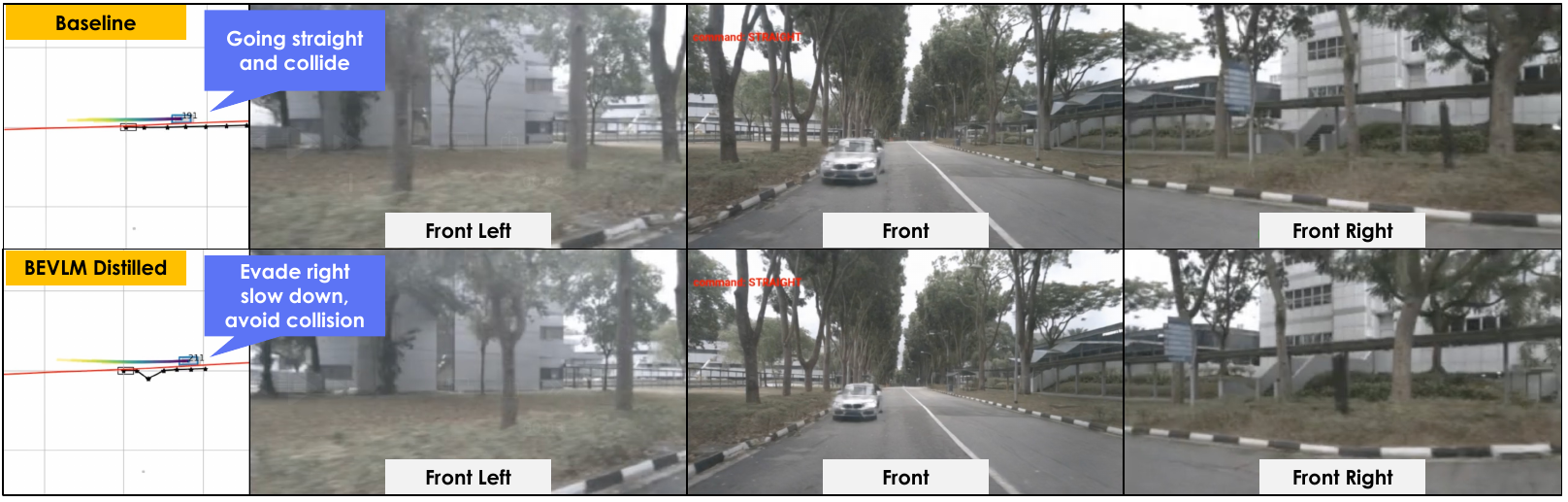}
        \caption{\textbf{Corner Case 4:} In this scenario, a white sedan is moving towards the ego vehicle in the same lane. To avoid the collision, the ego vehicle is supposed to evade by turning right.
        }
        \label{fig:neuroncap_4}
    \end{subfigure}
    \caption{\textbf{Qualitative NeuroNCAP Results.} Two additional closed-loop planning scenarios are presented for comparison between the baseline and distilled models. The distilled model demonstrates improved decision-making under safety-critical scenarios, successfully performing \textit{left evade} in corner case 3 and \textit{right evade} in corner case 4 to avoid potential collisions, where the baseline model fails.}
    \label{fig:neuroncap_supp}
\end{figure*}

\begin{figure*}
    \centering
    \includegraphics[width=\textwidth]{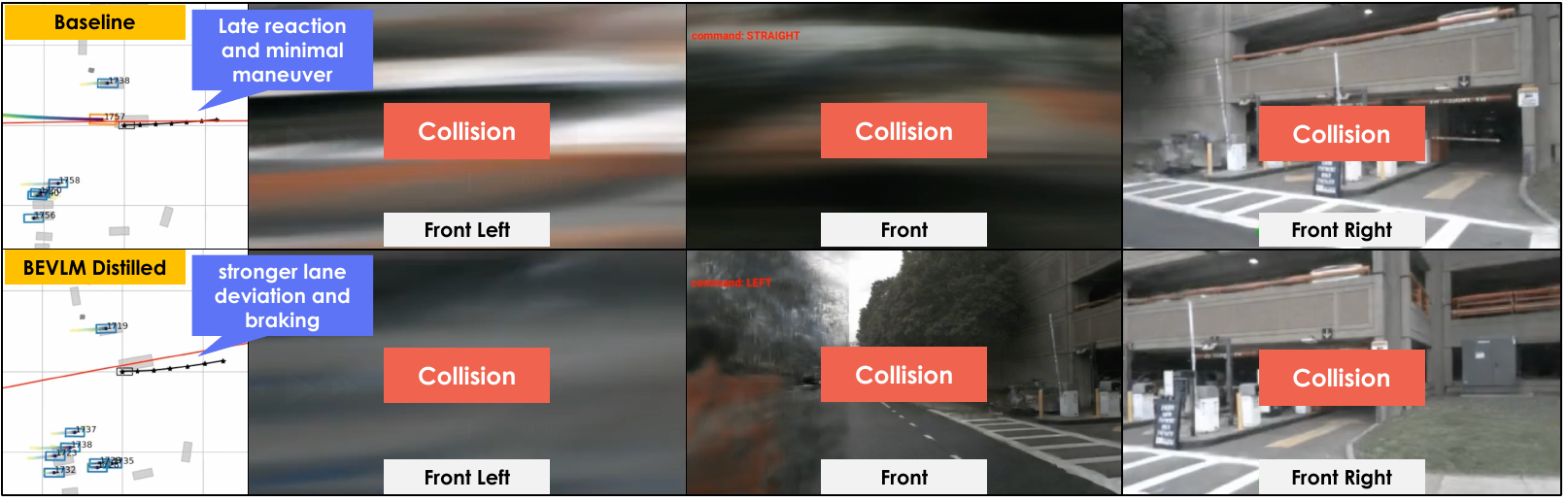}
    \label{fig:neuroncap_5}
    \caption{\textbf{Corner Case 5 (Failure Case).} Both the distilled model and the baseline model crash in this challenging scenario. However, our distilled model initiates the evasive maneuver much earlier, deviating more strongly from the lane center at impact time. As shown in the velocity profile in \cref{fig:velocity}, it also starts braking early, showing better safety awareness in this failure case.
    }
    \label{fig:neuroncap_failure}
\end{figure*}

\begin{figure*}[t]
    \centering
    \includegraphics[width=\linewidth]{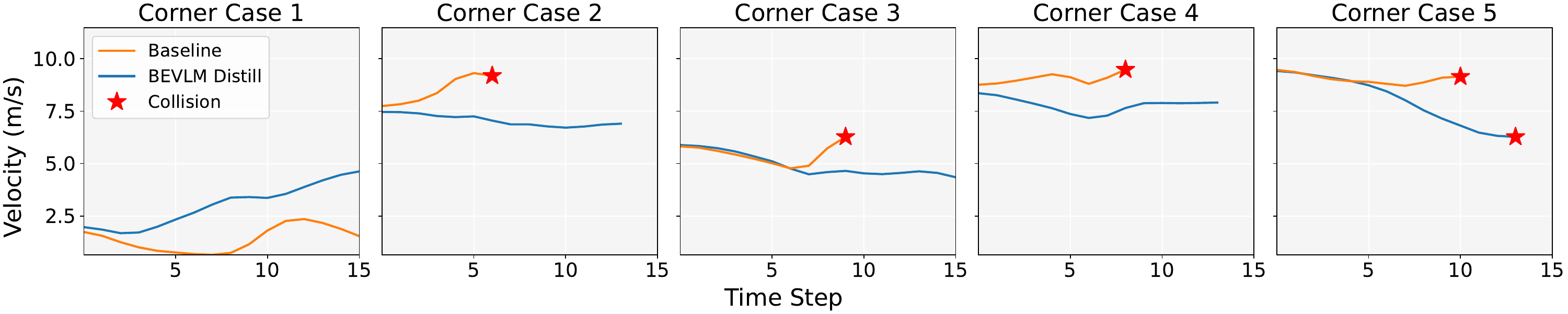}
    \caption{\textbf{Velocity Comparison.} We visualize the vehicle velocity for each corner case shown in the main paper and the Appendix. \textit{Case 1}: The distilled model anticipates blockage and turns swiftly to avoid the collision, while the baseline model proceeds hesitantly and collides. \textit{Case 2}: The distilled model turns right and avoids the collision, while the baseline model speeds up and collides. \textit{Case 3}: Speed is similar, but the distilled model turns left and successfully avoids the collision. \textit{Case 4}: The distilled model brakes and turns to avoid collision. \textit{Case 5 (Failure Case)}: Both models brake, but the distilled model brakes harder before the collision.
    }
    \label{fig:velocity}
\end{figure*}

\subsection{More Qualitative Results}

We present two more qualitative results in \cref{fig:neuroncap_supp}. In the third corner case, the ego vehicle is moving straight while a motorcycle cuts into the lane from the left. The baseline model tries to evade by turning right, which results in a collision since it overlaps with the future trajectory of the motorcycle. In contrast, the distilled model considers the moving trajectory of the motorcycle and evades by turning left, successfully avoiding the collision.

In the fourth corner case scenario, there is a white sedan moving towards the ego vehicle in the same lane. The baseline model reacts with minimal maneuver and crashes into the oncoming car. In contrast, our distilled model reacts by turning right and slowing down to evade and prevent the collision.

\subsection{Failure Cases of Distilled Model}
Corner case 5 in \cref{fig:neuroncap_failure} shows a failure case of the distilled model.
The challenging scenario requires evading an oncoming vehicle with an atypical appearance that drives on the wrong side of the road. The baseline UniAD model barely reacts and has a full frontal collision. Our distilled model cannot avoid the collision either, but it reacts earlier by braking and steering to the right, resulting in stronger lane deviation and lower collision impact. The contrast is particularly visible in the velocity profile in \cref{fig:velocity}, where the baseline keeps its velocity, while our distilled model reduces its velocity by around \SI{35}{\percent}, reducing the kinetic energy by more than half.

\subsection{Analysis of Impact Severity} \label{sec:impact_severity}

We also visualize the ego vehicle speed at each scenario in the main paper and in the Appendix, shown in \cref{fig:velocity}. The distilled model shows an obvious trend of slowing down under all the scenarios except case 1 (\ie, \cref{fig:neuroncap_1} in the main paper). For case 1, the distilled model anticipates the blockage and turns swiftly before another vehicle approaching from behind, while the baseline model proceeds hesitantly and eventually collides. For other cases, the baseline model shows little safety awareness and maintains the speed even before a collision. In the final cases, even when both models collide with the truck, the distilled model brakes harder before the collision, leading to a lower collision severity.
\section{Broader Impact and Limitations}
\label{supp:limitation}

In this section, we discuss the broader implications of our
study and acknowledge its potential limitations.

\subsection{Broader Impact}

We demonstrate the potential of BEV representation to enable better spatial reasoning in autonomous driving. While most of our evaluation focuses on end-to-end driving with conventional pipelines (\eg, UniAD~\cite{hu2023planning}), we hope this can provide insights into designing BEV-based VLAs that enhance the spatial reasoning capability while also benefiting from the semantic knowledge needed for reasoning in corner cases.

\subsection{Limitations}

In this subsection, we discuss the limitations of our work, which are mainly caused by the heavy computational costs. Additionally, we provide justification of why the main contribution of this work is still valid, given these limitations.

\subsubsection{BEV Model Architecture Generality.}
\label{sec:bev-arch}
We conducted our primary experiments using the BEV encoder architecture initially proposed in BEVFormer~\cite{li2022bevformer} and subsequently adopted by UniAD~\cite{hu2023planning} and VAD~\cite{jiang2023vad}. This choice is motivated by the fact that several representative end-to-end autonomous driving models are built upon this common BEV encoder architecture, often differing only in their task-specific decoder heads~\cite{jiang2023vad}. Moreover, our proposed framework does not constrain the BEV architecture design as long as the intermediate BEV representation can be generated. 

Meanwhile, we acknowledge the importance of demonstrating the generality of our proposed framework by experimenting with alternative BEV encoder architectures. However, conducting such experiments is constrained by the significant computational resources required. The semantic distillation step alone takes approximately 100 hours using 8 NVIDIA A100 80GB GPUs for the 8B LLMs. The subsequent end-to-end model training requires an additional 115 hours on the same hardware setup. Given this high computational cost, we must defer the exploration of other BEV architectures to future work, to demonstrate the broad applicability of our representation studies and the BEVLM framework.

\subsubsection{Human Labor-Free Distillation.}

We recognize the current limitation: the demonstrated safety improvement is primarily based on a human-curated dataset. The main focus of this paper is the rigorous comparison of visual representations and the introduction of the semantic distillation framework itself. We note that generating high-quality pseudo-labels is a non-trivial process, requiring careful question design, data filtering, and extensive inference resources from large-scale VLMs. Therefore, to prioritize the core framework's contribution, we regard exploring unsupervised, human-labor-free semantic distillation for future work.

\subsubsection{Further Scaling.}

In this work, we primarily conduct experiments with 1B/8B-scale LLMs on the DriveLM-nuScenes dataset. Future work can explore leveraging larger LLMs with more data for further scaling of the proposed approach.

\subsubsection{VLA Application.}

We mainly show the advantage of the distilled BEV representation on the traditional end-to-end models~\cite{hu2023planning, jiang2023vad}. Future work can explore if the distilled BEV representation can also benefit VLA for better spatial understanding while preserving the semantic richness.

\end{document}